\documentclass{l4dc2026}

\title[Adaptive Hierarchical RL-MPC]{Learning to Plan, Planning to Learn: Adaptive Hierarchical RL-MPC for Sample-Efficient Decision Making}
\usepackage{times}

\usepackage[dvipsnames]{xcolor}
\usepackage{float}
\usepackage{amsfonts}
\usepackage{fancyhdr}
\usepackage{comment}
\usepackage{amsmath}
\usepackage{changepage}
\usepackage{amssymb}
\usepackage{graphicx}
\usepackage{adjustbox}
\usepackage{latexsym}
\usepackage{enumerate}
\usepackage{float}
\usepackage{subcaption}
\usepackage{bm}

\usepackage[labelfont=bf,font=small]{caption}
\usepackage{mathtools}
\usepackage{bbm}
\usepackage{bm}
\usepackage{booktabs}
\usepackage{multirow}
\usepackage{makecell}
\usepackage{colortbl}
\usepackage{balance}

\usepackage{textcomp}
\usepackage{stfloats}  

\usepackage{fontawesome5}  

\newtheorem{assumption}{Assumption}    

\newtheorem*{proposition*}{Proposition}

\DeclareDocumentEnvironment{example}{}{\noindent\textbf{Running example:}\itshape}{}

\usepackage{ifthen}
\newboolean{include-notes}
\newboolean{include-new}
\newboolean{include-remove}
\setboolean{include-notes}{true}
\setboolean{include-new}{false}
\setboolean{include-remove}{false}

\usepackage[dvipsnames]{xcolor}
\usepackage[normalem]{ulem}

\newcommand{\princeton}[1]{\ifthenelse{\boolean{include-notes}}{\textcolor{orange}{#1}}{}}

\newcommand{\Span}{\mathrm{span}}

\DeclareMathOperator{\expectation}{\mathbb{E}}

\newcommand{\Env}{\operatorname{env}}
\newcommand{\Virt}{\operatorname{virt}}
\newcommand{\RL}{\operatorname{RL}}
\newcommand{\MPPI}{\operatorname{MPPI}}

\newcommand{\mdp}{\mathcal{M}}

\coltauthor{
 \Name{Toshiaki Hori} \Email{toshiaki.hori@tri.global}\\
 \Name{Jonathan DeCastro} \Email{jonathan.decastro@tri.global}\\
 \Name{Deepak Gopinath} \Email{deepak.gopinath@tri.global}\\
 \Name{Avinash Balachandran}\footnote{Los Altos, CA, 94022, USA} \Email{avinash.balachandran@tri.global}\\
 \Name{Guy Rosman} \Email{guy.rosman@tri.global}\\
 \addr Toyota Research Institute, Cambridge, MA 02139, USA
 \AND
}

\begin{document}

\maketitle

\begin{abstract}%
We propose a new approach for solving planning problems with a hierarchical structure, fusing reinforcement learning and MPC planning. Our formulation tightly and elegantly couples the two planning paradigms. It leverages reinforcement learning actions to inform the MPPI sampler, and adaptively aggregates MPPI samples to inform the value estimation. The resulting adaptive process leverages further MPPI exploration where value estimates are uncertain, and improves training robustness and the overall resulting policies.

This results in a robust planning approach that can handle complex planning problems and easily adapts to different applications, as demonstrated over several domains, including race driving, modified Acrobot, and Lunar Lander with added obstacles. Our results in these domains show better data efficiency and overall performance in terms of both rewards and task success, with up to a 72\% increase in success rate compared to existing approaches, as well as accelerated convergence ($\times 2.1$) compared to non-adaptive sampling. 
\end{abstract}

\section{Introduction}
Reinforcement learning (RL) has enjoyed widespread success in constructing control policies for embodied applications, including robotics~\cite{makoviychuk2021isaac}, fully- / partially-autonomous vehicles~\cite{kiran2021deep}, and board games~\cite{silver2017mastering}.  Nonetheless, learning RL-based controllers can be challenging when faced with environments and physical embodiments where it is costly or unsafe to interact with the target environment.
In domains where exploration data can be acquired in abundance, e.g., those with highly parallelizable simulators such as robotics, policies can be trained to solve very complex tasks.  However, in domains where data is more limited, e.g., domains with expensive-to-collect human data,  uncertain or unmodeled behaviors requiring more data to estimate, safety concerns limiting data collection, or those requiring extensive computation, more sample-efficient learning is needed.
In contrast to RL which seeks to solve for a globally-optimal policy, model-predictive control (MPC), and its sampling-based variants (e.g., MPPI), aim to solve a finite-horizon optimal control problem online by reasoning explicitly over local system behavior, usually using simplified models of the environment.  They are well-suited to optimizing for short-term objectives, though they are often extended with a terminal cost that can capture effects beyond the planning horizon. 
Numerous recent works have blended model-free RL and MPC to achieve data efficiency and safety using a variety of techniques~\cite{Romero2023ActorCriticMPC, AC4MPC, blending_mpc_value_approximation, info_model_predictive_q_learning, infusing_mpc_meta_rl, Hansen2024TD-MPC2}. 
However, such methods usually focus on specific cost formulations, couple RL and MPC in specific ways, or do not exploit RL and MPC problem structure. 

An underexplored idea involves treating an MPPI controller as a \textit{structured sampler}, which offers similar benefits to data augmentation -- 
it can cheaply generate plausible trajectories from an approximate model that can augment the RL training data.
The sampler's longer horizon affords more accurate value function learning, and the hierarchical decomposition offers clear locations to separate the role of the RL and MPC algorithms.  
Rather than optimizing each component separately through cost function coupling, we aim to explicitly reuse sampled MPPI trajectories as approximate, virtual rollouts to guide both value function learning and policy updates.  
Ultimately, drawing such samples reduces the number of real environment steps needed to learn a performant policy, without resorting to additional heuristics (e.g. reward shaping, curricula, feature selection).  This is crucial for settings where actual environment queries are extremely expensive or unsafe to collect.
Such a framework lends to a principled solution for balancing model-based supervision for achieving desired control performance and model-free exploration for achieving faster convergence and more efficient data utilization. 
The key challenge is that the distribution mismatch between the approximate model and the true environment, combined with off-policy biases from the MPPI sampler, can introduce systematic errors in value function estimates.

In this work, we propose an integrated RL–MPC framework that couples a high-level RL policy with a sample-based MPC controller via shared value function updates. Our contributions are threefold: (1) we introduce a bi-directional training pipeline that allows sample-based rollouts to improve value function learning as well as leverages improved policy estimates to tune the samples in a general setting, (2) we provide a means for adjusting the mixing of datasets and discuss how this maps to curriculum that initially explores safe, local rollouts, while eventually extending exploration to longer-horizon rollouts, and (3) we provide empirical evidence that coupling value-aware exploration with MPC-based rollouts significantly enhances sample efficiency and safety.

\section{Related Work}
Our work improves on the intersection of two main research threads in planning approaches. One thread is reinforcement learning \citep{SuttonBook,9904958}, more specifically, model-based reinforcement learning~\cite{MAL-086}. The other main thread is that of MPC techniques~\cite{schwenzer2021review}, and MPPI~\cite{info_theoretic_mpc_for_mbrl,Bhardwaj2020ITMPQ} based approaches within them.

Approaches for model-based reinforcement learning (MBRL) (e.g.,~\citep{Hafner2020Dreamer, Janner2019MBPO}) have demonstrated the benefit of learning latent world models that enable planning in imagination. However, such methods are costly to train, as a necessary condition for good policy performance is a well-trained model.  In contrast, hybrid RL–MPC methods~\citep{Nagabandi2018NNMPC, okada2020variational,li2025unifying,mundheda2025teacher} allow for more flexibility, and learned dynamics into MPC loops, but typically without a feedback mechanism from MPC to the RL updates (in the actor or critic).

With hierarchical RL showing promise for sample efficiency~\cite{wen2020efficiency}, recent work on blending MPC and value function learning have been proposed~\citep{DBLP:journals/corr/abs-1908-06012,Bhardwaj2020ITMPQ, blending_mpc_value_approximation, ac_mpc, AC4MPC,Serra-Gomez2025KL}.  Each focus on new strategies for blending, and the interface between the components for achieving better sample efficiency and to better balance exploration vs. modeling the environment.  

In~\cite{blending_mpc_value_approximation}, the authors aim to balance the estimates from a MPC-derived local Q-function with a RL-estimated value function.
Our approach differs by closing this loop; sampled MPC rollouts act as structured exploration priors that inform value function and policy updates to accelerate high-level learning.
Some approaches leveraged RL and MPC by leveraging RL to improve the MPC estimates~\cite{bootstrappedmpc}, or leverage MPPI samples to help robustify the RL training~\cite{mundheda2025teacher}, but not within a single coherent co-training framework.
Several works have also looked at specific complex domains where RL and MPC can be combined, along with domain-specific insights~\cite{nguyen2025td,kotecha2025realtimegaitadaptationquadrupeds}.

A related thread of research focuses on safe RL~\citep{Berkenkamp2017SafeRL, garcia2015comprehensive}, where constraints are enforced via Lyapunov conditions or shielding mechanisms. Our approach offers an alternative safety-critical reasoning to the MPC layer while using RL to shape high-level intent. In human-interactive domains, several works (e.g.~\cite{Sadigh2018PlanningHuman, Rudenko2020}) address prediction and intent inference; the proposed RL–MPC coupling provides a mechanism to integrate such predictive models into closed-loop decision-making.

\section{Background and Problem Statement}
We assume access to an MDP $\mdp = \langle \mathcal{S}, \mathcal{A}, r, P, \gamma, \mu \rangle$, where $s_t \in \mathcal{S}$ is the state/observation at discrete time $t$,  $a_t \in \mathcal{A}$ be the action taken at $t$, $r: \mathcal{S} \times \mathcal{A} \mapsto \mathbb{R}$ is a reward function, $P: \mathcal{S} \times \mathcal{A} \mapsto \Delta(\mathcal{S})$ is a transition probability, $\gamma \in [0, 1]$ is a discount factor, and $\mu \in \Delta(\mathcal{S})$ is the initial state distribution.  The general RL objective is to find an optimal policy $\pi^{\star}_{\phi}$ (under parameters $\phi$) that maximize the infinite-time discounted return
\begin{equation}
    \pi^{\star}_{\phi} = {\arg\max}_\pi\expectation_{\mathcal{M}, \pi, s_0 \sim \mu}\left[\sum_{\tau=t}^{\infty} \gamma^\tau r(s_\tau, a_\tau) \right].
\end{equation}

Model-predictive path integral (MPPI,~\cite{info_theoretic_mpc_for_mbrl}) is a sample-based algorithm that solves an optimal control problem expressed by a cost function and a model $\hat{\mathcal{M}} = \langle \mathcal{X}, \mathcal{U}, \hat{r}, \hat{P}, H, \hat\mu \rangle$  that \emph{approximates} $\mdp$, where $x_t\in\mathcal{X}$ is the system state, $u_t\in\mathcal{U}$ is the control input, and $\hat{P}$ is the dynamical system $\hat{P}: \mathcal{X} \times \mathcal{U} \mapsto \Delta(\mathcal{X})$ (e.g.~equations of motion, a neural approximator, etc.) $\hat{r}: \mathcal{X} \times \mathcal{U} \mapsto \mathbb{R}$ is an approximate reward function, $H > 0$ is a integer-valued finite planning horizon, and $\hat{\mu} \in \Delta(\mathcal{X})$ is an approximation to $\mu$.  We assume the existence of a state interface map $\psi : \mathcal{X} \mapsto \mathcal{S}$ that converts MPPI states to RL observations (e.g., projecting a full physical state to a lower-dimensional observation vector).

MPPI draws $N$ input sequences around a nominal $u_{t:t+H-1}$, then samples additive control noise according to $\varepsilon_\tau^{i} \sim \mathcal{N}(0,\Sigma)$. The objective evaluates, for each sample $k\in 1,\ldots, K$, the cost
\begin{equation}
\label{eq:mmpi_cost}
    J_k^{\hat{P}}\left(u_{t:t+H-1}, a_t\right) = \expectation_{\hat P}\left[ \sum_{\tau = t}^{t + H - 1} J\left(\hat{x}^k_\tau, u^k_\tau + \varepsilon_\tau^{k}; a_t\right) + \phi\left(\hat{x}^k_{t + H}; a_t\right)
    \right]
\end{equation}
to be minimized.  The $k$-th state is updated according to $\hat x_{\tau+1}^{k} \in \hat{P}(\hat x_\tau^{k},u_\tau^{k} + \varepsilon_\tau^{k})$, with $ \hat x_t^{k} = x_t$.
The MPPI update rule for the \(k\)-th candidate is defined by:
\begin{equation}
\label{eq:mppi_update_law}
    u_t \leftarrow u_t + \sum_{j=1}^{K} \tilde w_{j}\,\varepsilon_{t}^{j},
    \qquad
    w_{k} \propto \exp\bigl(-J_{k}^{\hat{P}}/\lambda_{\mathrm{temp}}\bigr),
    \qquad
    \tilde w_{k}=\frac{w_{k}}{\sum_{j=1}^{K} w_{j}},
\end{equation}
where $\lambda_{\mathrm{temp}} > 0$ is a temperature parameter.

In~\eqref{eq:mmpi_cost}, we allow the running cost $J(\cdot, \cdot; a_t)$ and terminal cost $\phi(\cdot; a_t)$ (accumulating costs at all times $t \ge H$) to couple hierarchically via the RL action $a_t$.  These costs take general forms in $a_\tau$, e.g. parametric forms~\cite{ac_mpc}, terminal value~\cite{Hansen2024TD-MPC2}, or target tracking~\cite{Wang2023LearningReferencesOnlineMPC}.  Note that these are defined independently of the approximate reward $\hat{r}$, due to any domain-specific adjustments when applying state-action tuples to the experience buffer.



\begin{figure}
    \centering
    \includegraphics[width=0.9\columnwidth]{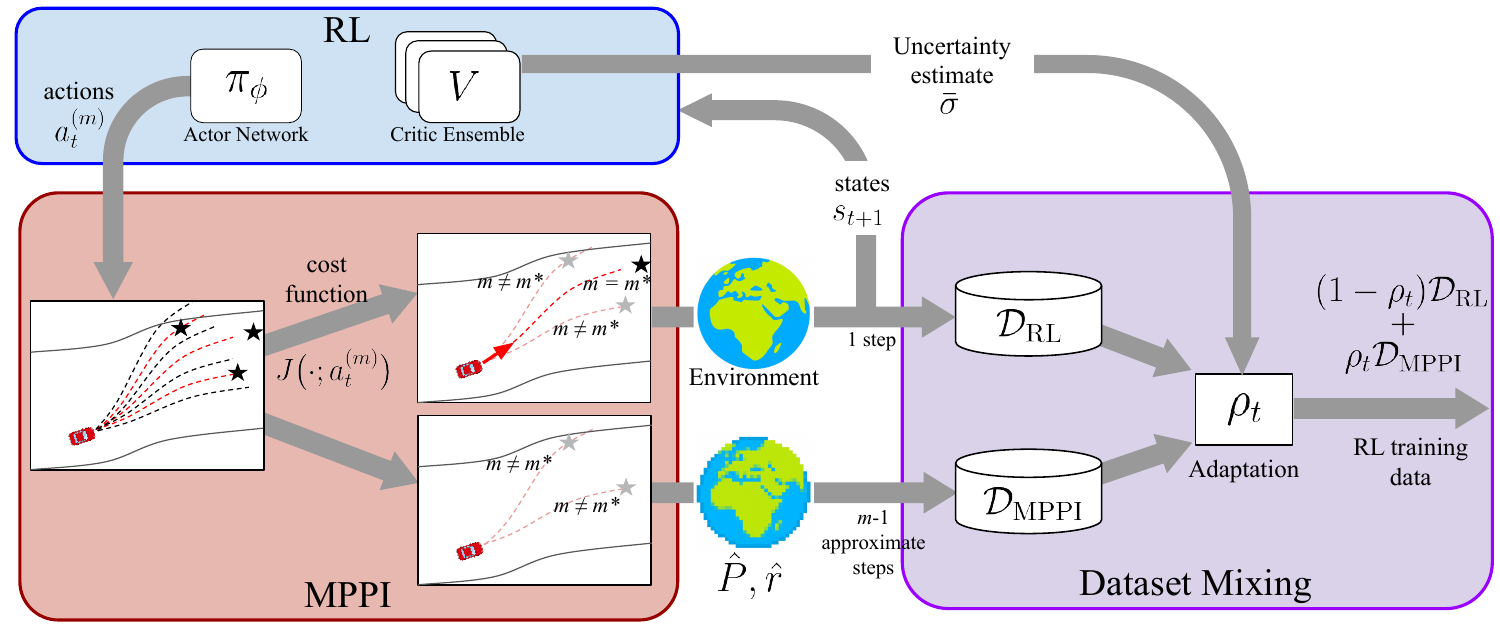}
    \caption{Diagram of the combined approach.  
    Samples from the RL policy generates actions that are fed to MPPI, which then generates a set of candidates $m_0, m_1, \ldots$.  One $m^{\star}$ is selected and applied to the real environment.  The remainder are stored in a buffer $\mathcal{D}_{\MPPI}$.  
    Data from the two buffers is sampled from a convex combination on the parameter $\rho_t$, defined by the uncertainty as estimated by a critic ensemble.  The data is passed to RL for value and policy iteration.}
    \label{fig:proposed_method_diagram}
\end{figure}

\subsection{Problem Statement}
Consider the case where we are given an MDP $\mdp$ describing a physical system embedded in an environment, but also have access to a surrogate $\hat\mdp$, which is used for MPPI and assumed fast to sample from, but bounded according to Assumption~\ref{assump:boundedness}.  
\begin{assumption}
\label{assump:boundedness}
    We assume the error between the approximate and true transition function is bounded according to $\alpha_P := \sup_{s, a, x, u} \left\| P(\cdot \mid s, a) - \hat{P}(\cdot \mid x, u)\right\|_{TV}$, where $\left\|\cdot\right\|_{TV}$ is the total variation norm.  We further assume the error between the surrogate and true reward functions is bounded according to $\alpha_r := \sup_{s, a, x, u} \left| r(s, a) - \hat{r}(x, u)\right|$.
\end{assumption}

Our aim is to unify RL and MPC in a way that improves sample efficiency over existing RL and MPC combinations, while eliminating bias due to the distribution shift when sampling from $\hat{\mathcal{M}}$.  We assume domains with reasonable complexity, that is, those exhibiting a combination of (hard) physical and (soft) task-related constraints with non-trivial transition functions (i.e. vehicle, aircraft, robot dynamics, etc.).

\section{Technical Approach}
We adopt a two-layer design, as shown in Fig.~\ref{fig:proposed_method_diagram}, where we assume an RL policy that outputs \emph{high-level tactical parameters} used by an MPPI cost function (e.g., target states, cost weights that penalize all state and action features, etc.), while an MPPI planning algorithm performs \emph{low-level operational control} on the system.  We posit that sampled rollouts from MPPI, which we call \emph{virtual rollouts}, can offer benefits as an additional, structured data source for value function learning.  This, in turn, offers overall benefits to learning a high-level RL policy.  Additionally, we posit that the learned policy can be used to sample a diverse set of behaviors that beneft MPPI.  Hence, the focus of this paper is in \emph{value-based} RL methods.

We propose a commonly-used hierarchical controller setup where, at the high level, a value-based RL scheme outputs a set of actions while, at the low level, MPPI computes a set of locally optimal low-level input sequences for each high-level action.  Of the candidate solutions, one optimal solution is applied to the environment; the remaining solutions are scored by a reward model and stored in an MPPI buffer. Crucially, for each high-level action we store only MPPI-feasible trajectories (not arbitrary samples) in an isolated buffer, so that the buffer reflects the controller’s optimality and keeps the data effectively on-policy with respect to the hierarchical controller.  

At each RL training iteration, we mix RL buffer data with MPPI buffer data, and use this augmented data to update the value function and RL policy.  Biases can be incurred in value learning due to any mismatches in the approximate model that MPPI uses for runtime optimization and any off-policy biases incurred due to the MPPI sampler.  To allow our approach to have zero bias with respect to modeling approximations, we propose a tunable influence ratio to allow the algorithm to control for uncertainty, and hence recover zero asymptotic bias.


\subsection{Coupling RL and MPPI}
Several value-based RL approaches have seen widespread use, including proximal policy optimization (PPO), soft actor-critic (SAC), Advantage Actor Critic (A2C), etc.  We choose PPO because the critic target considers future rewards, and hence our trajectory sampling scheme is amenable to that setup.  On the other hand, common SAC algorithms only use a one-step lookahead as a critic target. 
The overall objective for both approaches is a composite of an actor loss, a critic loss and an entropy loss to encourage exploration.  
PPO is an \emph{on-policy} algorithm, meaning that experience is accumulated using the policy at the current training iteration.  In PPO's experience buffer, we can extend and diversify future policy behavior by augmenting with MPPI rollouts without executing them on the actual environment.  
\paragraph{Procedure}
Our algorithm is presented in Alg.~\ref{alg:rl_mppi}.  At each time \(t\) we draw \(M\) high-level candidate actions 
\begin{equation}
\label{eq:rl_action_sampling}
    a_t^{(m)} \sim \pi_{\phi}(\cdot \mid s_t), \qquad m=1,\dots,M.
\end{equation}
We then sample a \emph{single} set of \(K\) rollout perturbation sequences
\(\{\varepsilon_{t:t+H-1}^{k}\}_{k=1}^{K}\) and, from it, generate the state-action trajectories $(\hat x_{t:t+H}^{k}, u_{t:t+H-1}^{k})$ using our approximate MPPI dynamics $\hat x_{\tau+1}^{k} \in \hat P\bigl(\hat x_{\tau}^{k},\,u_{\tau}^{k}\bigr)$, with $\hat x_t^{k}=x_t$.
For each $m$, we apply the cost $J_k^{\hat{P}}\left(u_{t:t+H-1}, a_t^{(m)}\right)$ in~\eqref{eq:mmpi_cost}, which is conditioned on the RL action $a_t^{(m)}$ that, when coupled with cost in this way, serves as a high-level knob that reshapes the MPPI objective.  For each $m$, we execute the control update step in~\eqref{eq:mppi_update_law}.  Sampling actions in~\eqref{eq:rl_action_sampling} in this way allows for a diverse set of feasible rollout options to enter the replay buffer.

\begin{algorithm2e}[t]
\small
\DontPrintSemicolon
\caption{RL$\,+\,$MPPI}\label{alg:rl_mppi}
\SetKwInput{Input}{Input}
\SetKwInput{Parameter}{Parameters}
\Input{dynamics model $\hat P$, reward model $\hat r$, policy $\pi_\theta$,
       value heads $\{V_d\}_{d=1}^D$,
       buffers $\mathcal{D}_{\RL},\mathcal{D}_{\MPPI}$}
\Parameter{episodes $N$, steps $T$, horizon $H$,
           MPPI candidates per action $M$, minibatch $B$,
           update period $N_{\text{upd}}$, EMA $\lambda$, init $\rho_0$}
\For{$i=1$ \KwTo $N$}{
  \For{$t=1$ \KwTo $T$}{
    $(a_t^{(1)},\ldots,a_t^{(m)}) \leftarrow \pi_\theta(\cdot\vert s_t)$\;
    $\{u_{t:t+H}^{(m)}\}_{m=1}^M, \{s_{t:t+H}^{(m)}\}_{m=1}^M \leftarrow$ MPPI\;
    \For{$m=1,\ldots,M$}{
      \eIf{$m=m^\star$}{
        $\mathcal{D}_{\RL} \leftarrow$ 1-step data\;
      }{
        $V_{target}(s_t) \leftarrow$ GAE value target~\cite{schulman2015high}\;
        $\mathcal{D}_{\MPPI} \leftarrow$ 1-step data includes $V_{target}(s_t)$\;
      }
    }
  }
  \If{$i \bmod N_{\textnormal{upd}}=0$}{
    Sample minibatches $\mathcal{B}_{\RL}^{(B)}\!\subset\!\mathcal{D}_{\RL}$, $\mathcal{B}_{\MPPI}^{(B)}\!\subset\!\mathcal{D}_{\MPPI}$\;
    Calculate $\rho_t$ via Alg.~\ref{alg:influ_ratio}\;
    $L \leftarrow (1-\rho_t) L_{\RL} + \rho_t L_{\MPPI}$\;
    Update $\theta \leftarrow \theta - \alpha \nabla_\theta L$\;
  }
}
\KwRet{$\theta$ (or best on validation)}\;
\end{algorithm2e}

\paragraph{Selected candidate $m^{\star}$}
We select one candidate $m^\star$ among $m=1,\dots,M$ to apply to the environment.  Although it is possible to optimally select among available candidates, in order not to introduce selection bias on our high-level policy $\pi_{\phi}$, we sample $m$ uniformly from the set, then apply it to the real environment.
We store the resulting tuple in the environment buffer, $\mathcal{D}_{\RL}$:
\begin{equation}
  \Big(s_t, a_t^{(m^\star)}, u_t^{(m^\star)}, r_t^{(m^\star)}, s_{t+1}^{(m^\star)}, \log\pi_{\mathrm{old}}(a_t^{(m^\star)}\!\mid s_t)\Big).
\end{equation}
Here, $\pi_{\mathrm{old}}$ denotes the PPO policy used to generate the sampled high-level action before the update.

\paragraph{Sampled candidates $m$}
We re-score samples from the remaining candidates $m\neq m^\star$ before storing to a buffer.  To diversify the buffer, we do not include the samples for the selected candidate.  As confirmed by experimentation, including $m^\star$ did not alter the results significantly.  We first define virtual transitions using a reward approximator $\hat r(x,u)$ and MPPI dynamics model $\hat P$.  Next, we assume the existence of an state interface map, $\psi: \mathcal{X} \mapsto \mathcal{S}$,
which allows conversion from MPPI states to RL states (e.g. mapping a set of physical states to a set of lower-dimensional observations).  We then use this map to accumulate samples 
\begin{align}
  \hat r_t^{(m)} = \hat r\big(x_t,u_t^{(m)}\big), \qquad
  \hat s_{t+1}^{(m)} = \psi\left( \hat P \big(x_t,u_t^{(m)}\big) \right).
\end{align}
Finally, we store in an MPPI buffer, $\mathcal{D}_{\MPPI}$, tuples of the form
\begin{equation}
  \Big(s_t,\ a_t^{(m)},\ u_t^{(m)},\ \hat r_t^{(m)},\ \hat s_{t+1}^{(m)},\ \log\pi_{\mathrm{old}}(a_t^{(m)}\!\mid s_t)\Big).
\end{equation}

\subsection{Dataset Mixing}
Given that our data are maintained in two buffers, we can utilize both for sampling training batches as well as for defining a composite loss function, with each component derived from one of the buffers. In the limit of large sample sizes, these two formulations are equivalent.

Let $\rho \in [0, 1]$ be a sample-based influence ratio between a buffer of real data $\mathcal{D}_{\RL}$ and virtual data $\mathcal{D}_{\MPPI}$.  At each step $t$, we sample $z_t \sim \mathrm{Bernoulli}(\rho)$ such that
\begin{equation}
\label{eq:sample_mixing}
    d_t = 
    \begin{cases}
    d_{\Env} \sim \mathcal{D}_{\RL} & z_t=0 \\
    d_{\Virt} \sim \mathcal{D}_{\MPPI} & z_t=1
    \end{cases}
\end{equation}
This induces a unified replay buffer with convex sampling distribution
\begin{equation}
  s,a \sim (1-\rho)\mathcal{D}_{\RL} + \rho \mathcal{D}_{\MPPI}.
\end{equation}
The equivalent PPO loss can be expressed as:
\begin{equation}
\label{eq:loss_weighting}
  L_{\mathrm{PPO}}(\phi)
  =
  (1-\rho) \expectation_{d_t \sim \mathcal{D}_{\RL}} \left[ L_{\mathrm{PPO}}(\phi; d_t) \right]
  +
  \rho \expectation_{d_t \sim \mathcal{D}_{\MPPI}} \left[ L_{\mathrm{PPO}}(\phi; d_t) \right].
\end{equation}
In Alg.~\ref{alg:rl_mppi}, we employ loss function weighting of~\eqref{eq:loss_weighting}.  Experiments in the Supplement confirm that the direct mixing scheme of~\eqref{eq:sample_mixing} is largely consistent with the loss-based weighting in~\eqref{eq:loss_weighting}.

As opposed to the current PPO practice of filling a buffer of on-policy data before invoking state updates, our scheme allows for a more frequent updating process.  Consider the case where policy updates occur after an experience buffer of $N_B$ is filled.  In standard PPO, this requires $N_B$ environment steps.  In our approach, assuming $M$ MPPI samples are obtained per RL step, mixing virtual rollouts can allow for an effective reduction to $N_B / M$ environment steps.  

\subsection{Bias in RL-MPC}
Under the proposed approach, we bound the gap between the optimal value $V_{\pi^{\star}, \mdp}$ and the estimated value $\hat{V}_{\pi, \mdp_{\rho}}$ of the idealized compositional policy under the mixed MDP, measuring the optimality under the approach, with proof in Appendix~\ref{sec:proofs}.

\begin{theorem}
\label{theorem:value_error}
Let $\pi$ be the policy of the combined RL-MPC approach and $\pi^{\star}$ be the optimal policy under the true MDP, and take the bounds from Assumption~\ref{assump:boundedness}.  Let $\hat{V}$ be a value function estimate.
We can bound the value function error using the proposed approach to the value obtained from the optimal policy in the true domain according to:
\begin{equation}
    \begin{aligned}
        \left\| V_{\pi^*, \mdp}(s) - \hat{V}_{\pi, \mdp_{\rho}}(s) \right\|_\infty &\le
        \rho \alpha_P H \gamma \frac{1-\gamma^{H-1}}{1-\gamma}\frac{\Span(r)}{2}
        + 
        \rho \alpha_r \frac{1-\gamma^{H}}{1-\gamma} \\
        & 
        +
        \gamma^H\varepsilon_V
        +
        \frac{4 R_{\max} \left(1 - \gamma^H\right)}{\left(1 - \gamma\right)^2} \cdot \max_s D_{TV}\left(\pi^{\star}(\cdot\mid s) \Vert \pi(\cdot\mid s)\right)
    \end{aligned}
\end{equation}
where $\Span(r) := \sup(r) - \inf(r)$, $\varepsilon_V := \sup_s\left|V(s) - \hat{V}(s)\right|$ and $R_{\max} := \sup_{s, a} \left| r(s, a) \right|$.
\end{theorem}
The result highlights three separable contributors to error: model mismatch due to $\mdp_{\rho}$, value approximation error from the terminal bootstrap, and a policy sub-optimality gap.  The first vanishes via appropriate adaptation of $\rho$, while the remaining terms vanish with training.  Moreover, as discussed in Corollary~\ref{cor:implementation_bound} (Appendix~\ref{sec:proofs}), finite MPPI samples introduce an additional resampling bias that does not decay with $\rho$; this further motivates annealing $\rho$ to reduce the weight placed on MPPI-derived estimates.

\subsection{Adaptive Influence Ratio}
While virtual data can improve exploration and data efficiency, the model terms in Theorem~\ref{theorem:value_error} 
scale linearly with the influence ratio $\rho$, so we wish to drive $\rho$ to zero.
We adapt the influence ratio dynamically in Alg.~\ref{alg:influ_ratio} by using ensemble disagreement as a proxy for when the critic has learned enough that virtual data is no longer needed.


Similar to~\cite{chen2017ucb, peer2021ensemble}, we estimate value uncertainty $\bar{\sigma}$ from an ensemble of value functions and construct a bounded confidence score $\omega_t=\tfrac{1}{1+\bar{\sigma}_t^2}$. This score is smoothed via an exponential moving average (EMA) when applied to $\rho$, so that, as value-function uncertainty decreases, $\rho_t$ is annealed toward zero.  The monotonic decay ensures that the algorithm eventually relies entirely on real data for the model error terms in Theorem~\ref{theorem:value_error} to vanish.
The smoothing parameter $\lambda$ controls the decay rate: larger $\lambda \rightarrow 1$ yields slower, stable annealing that filters out transient fluctuations in ensemble variance, while smaller $\lambda \rightarrow 0$ makes $\rho$ more responsive.  The initial value $\rho_0$ sets the level of trust in the virtual data during early training.  Higher $\rho_0$ encourages more aggressive use of MPPI rollouts when real data is scarce, at the cost of initial bias if the surrogate model is poor (Table~\ref{tab:rho_comparison_result} captures this tradeoff).


\begin{algorithm2e}[t]
  \small
  \DontPrintSemicolon
  \caption{Online weight update from ensemble uncertainty}
  \label{alg:influ_ratio}
  \SetKwInput{Input}{Input}
  \SetKwInput{Parameter}{Parameters}
  \Input{$V\in\mathbb{R}^{B\times D}$ (batch $\times$ ensemble), previous EMA $\Omega_t\!\ge0$, previous weight $\rho_t\!\ge0$}
  \Parameter{ensemble size $D$, batch size $B$, smoothing factor $\lambda\in[0,1)$, $\rho_0$, $\Omega_0$}
  $\mu_b \gets \frac{1}{D}\sum_{d=1}^{D} V_{:,d}$ \tcp*{per-sample mean across ensemble}
  $\sigma^{2}_b \gets \frac{1}{D}\sum_{d=1}^{D}\!\bigl(V_{:,d}-\mu_b)^2$ \tcp*{per-sample ensemble variance ($\in\mathbb{R}^{B}$)}
  $\bar{\sigma}^{2} \gets \frac{1}{B}\sum_{b=1}^{B} \sigma^{2}_b$ \tcp*{aggregate uncertainty over batch}
  $\omega \gets \frac{1}{1+\bar{\sigma}^{2}}$\tcp*{confidence score}
  $\Omega_{t+1} \gets \lambda\,\Omega_t + (1-\lambda)\,\omega$\tcp*{EMA}
  $\eta \gets (1-\lambda)\,\Omega_{t+1}$\tcp*{decay rate}
  $\rho_{t+1} \gets \max\!\bigl(0,\ \rho_t (1-\eta)\bigr)$\tcp*{non-negativity clamp}
  \KwRet{$\Omega_{t+1},\ \rho_{t+1}$}
\end{algorithm2e}

\section{Experiments and Results}
\label{sec:result}

We evaluate control performance in three environments:
(i) Acrobot~\citep{NIPS1995_8f1d4362}, (ii) Lunar Lander~\citep{brockman2016openai}, and (iii) Racing (CARLA)~\citep{decastro2025dreaming}.
In Acrobot, the environment dynamics and reward are intentionally simple, and the dynamics/reward models used by our method match the ground-truth environment exactly.
In Lunar Lander, which is built on Box2D physics, our method uses approximate dynamics and reward models.
In Racing (CARLA), we adopt an even coarser (more misspecified) dynamics and reward model.
This setup creates a spectrum of model mismatch—from exact (Acrobot) to approximate (Lunar Lander) to highly-mismatched (Racing)—allowing us to assess robustness.
Specifically, the RL action injected into MPPI is defined as target angles in Acrobot, and as target velocities in Lunar Lander and Racing.

Across all environments, we introduce a \emph{Danger Zone} to increase task difficulty:
these are randomly-generate unsafe regions that yield a large negative reward upon entry.
See the supplement for further details (e.g., state / action spaces, rewards, and tasks).

We structure the analysis in two stages. First, we compare four methods to highlight inter-method diffenences: (i) PPO (baseline), (ii) SAC (baseline), (iii) PPO-MPPI with no virtual data ($\rho=0.0$), and (iv) PPO-MPPI with a fixed virtual-data mixing ratio ($\rho=0.3$). Second, within the proposed method, we study the effect of the mixing ratio by varying $\rho$ across fixed values and by using the adaptation scheme defined in Algorithm ~\ref{alg:influ_ratio}. This two-step design isolates the contributions of planning and virtual–data mixing, allowing us to examine how the choice of $\rho$ influences outcomes under different levels of model mismatch.

\subsection{Comparison Result of Methods}
We observe a consistent trend across three environments (Table~\ref{tab:method_comparison_result}). The baseline \textbf{PPO} and \textbf{SAC} frequently explores within the Danger Zone during early training and subsequently collapses to behaviors that remain near the initial state. As a result, task success---measured by success/finish rates---remains low, indicating entrapment in local optimal. 
Augmenting with \textbf{MPPI} (\(\rho=0.0\)) allows us to introduce soft constraints in the planner, leading from the outset to exploratory trajectories that repeatedly avoid the Danger Zone; in Acrobot, this yields a \(100\%\) success rate.  However, in the remaining environments, success rates drop dramatically.
By contrast, applying virtual-data mixing at \(\rho=0.3\) allows for exploration along the contour and within the interior of the Danger Zones.  
The results confirm that collecting multiple samples per step increases coverage of boundary states, alleviates data imbalance, and leads to higher success/finish rates than the other methods.

\begin{figure}
  \centering
  \begin{minipage}[t]{0.3\textwidth}\centering
    \includegraphics[height=0.11\textheight]{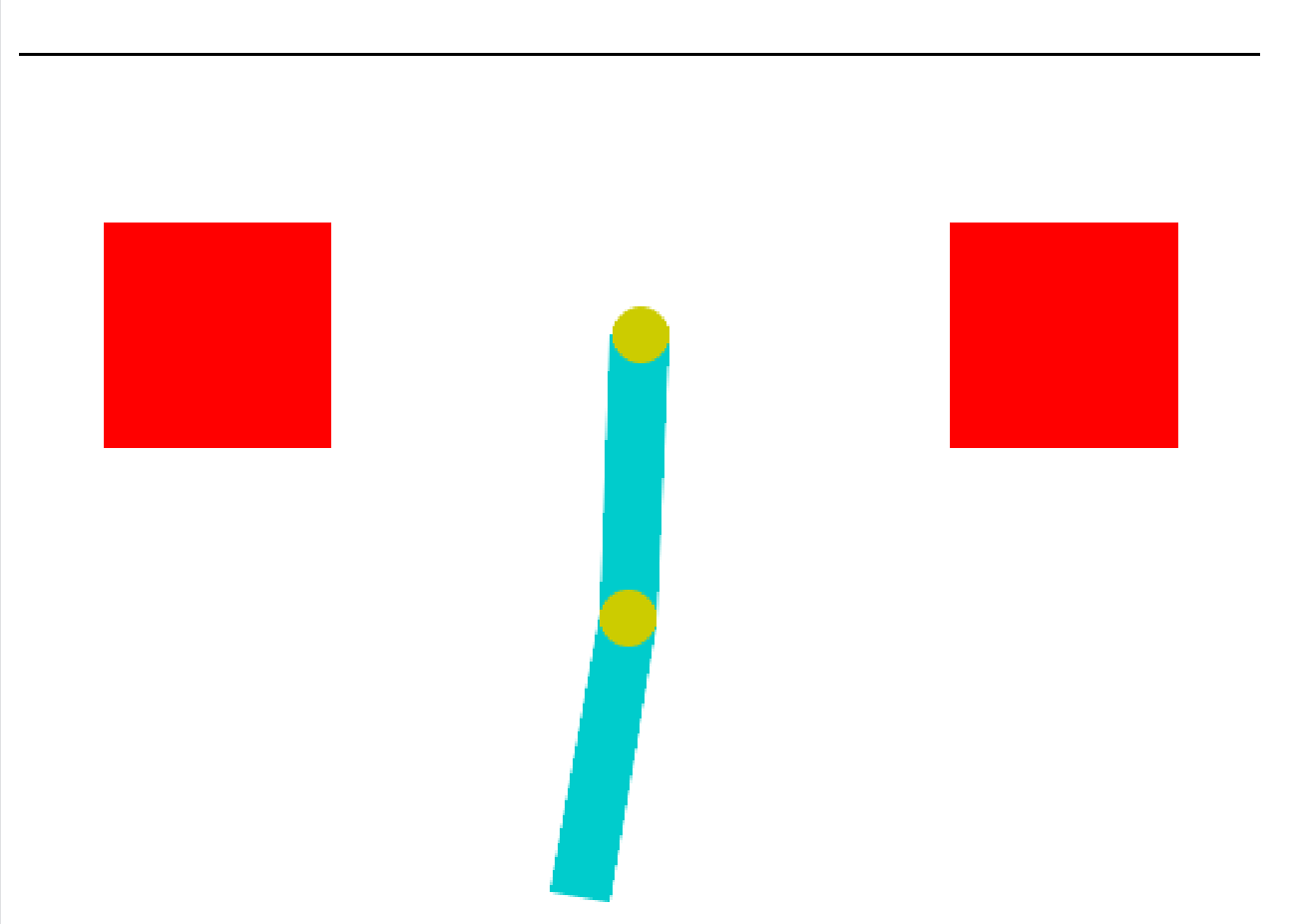}\\
    \small (a) Acrobot Environment
  \end{minipage}\hfill
  \begin{minipage}[t]{0.3\textwidth}\centering
    \includegraphics[height=0.11\textheight]{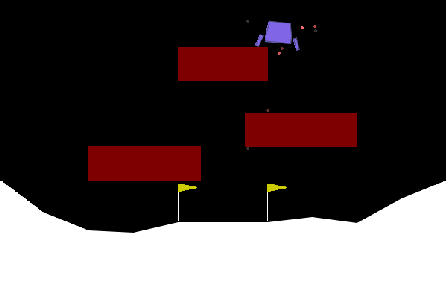}\\
    \small (b) Lunar Lander Environment
  \end{minipage}\hfill
  \begin{minipage}[t]{0.3\textwidth}\centering
    \includegraphics[height=0.11\textheight]{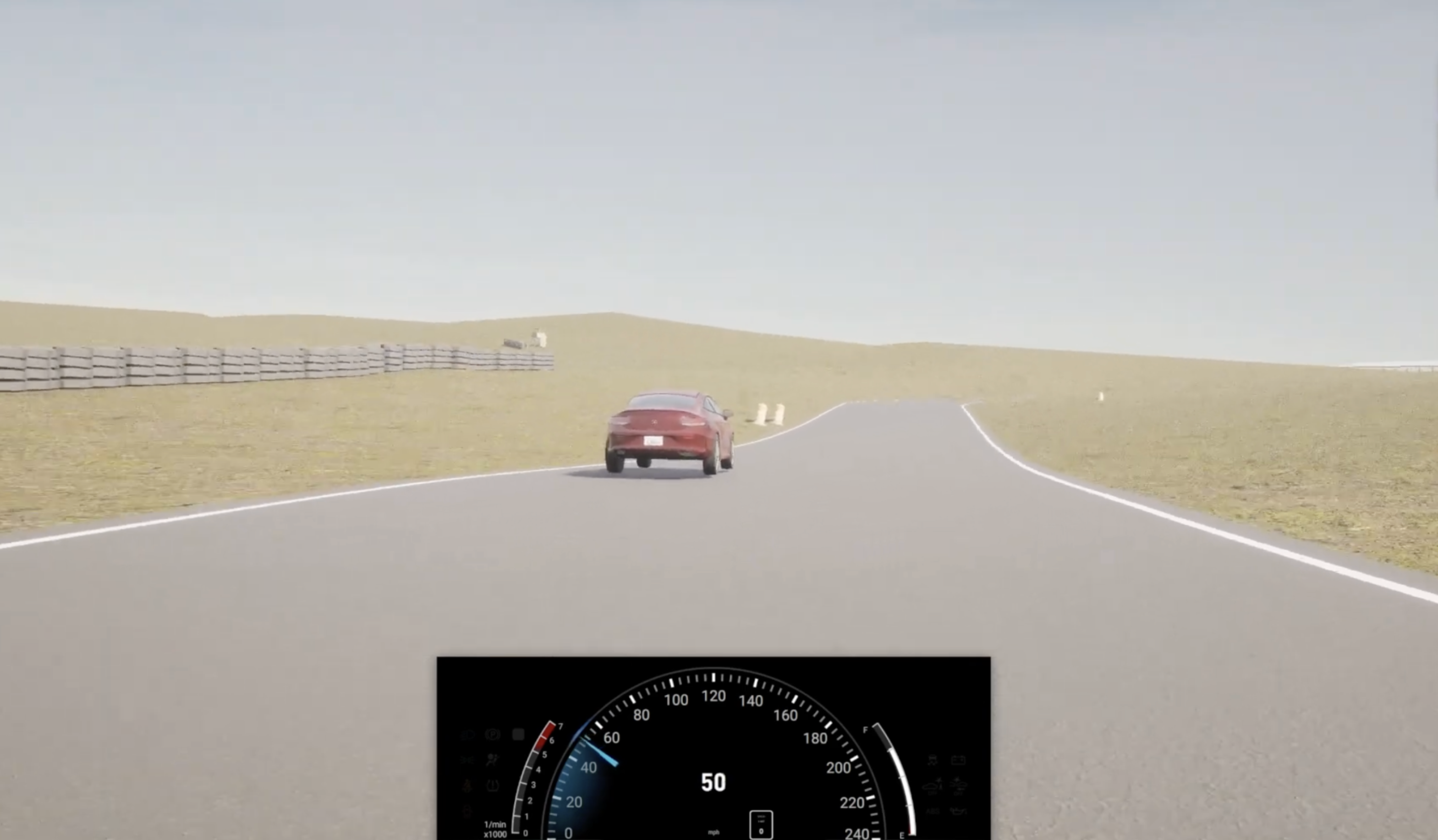}\\
    \small (c) CARLA racing Environment
  \end{minipage}\hfill
  \caption{Experimental environments.}\label{fig:exp_envs}
\end{figure}

\begin{table}[t]
  \centering
  \tiny
  \setlength{\tabcolsep}{2pt}
  \renewcommand{\arraystretch}{1.20}
  \caption{Evaluation across four methods. Entries are \textit{mean} $\pm$ \textit{std} over 50 episodes per each method. $\uparrow$ / $\downarrow$ indicate whether higher / lower equates to better performance, with \textbf{bold} being best.}
  \begin{tabular}{c c c c c c c c c c}
    \toprule
    & \multicolumn{3}{c}{\textbf{Acrobot}}
    & \multicolumn{3}{c}{\textbf{Lunar Lander}}
    & \multicolumn{3}{c}{\textbf{Racing}} \\
    \cline{2-4}\cline{5-7}\cline{8-10}
    \textbf{Method}
    & \textbf{Success [\%]} & \textbf{Step [$\downarrow$]} & \textbf{Reward[$\uparrow$]}
    & \textbf{Success [\%]} & \makecell{\textbf{Dist. to}\\\textbf{Goal [$\downarrow$]}} & \textbf{Reward[$\uparrow$]}
    & \textbf{Finish [\%]} & \makecell{\textbf{Coll./}\\\textbf{Off-track [$\downarrow$]}} & \textbf{Reward[$\uparrow$]} \\
    \midrule
    \makecell[c]{\textbf{PPO}}
    & \makecell[c]{$0.40$\\$\pm\,0.49$}
    & \makecell[c]{$332.6$\\$\pm\,205.1$}
    & \makecell[c]{$-347.1$\\$\pm\,188.4$}
    & \makecell[c]{$0.00$\\$\pm\,0.00$}
    & \makecell[c]{$1.30$\\$\pm\,0.35$}
    & \makecell[c]{$-141.9$\\$\pm\,84.9$}
    & \makecell[c]{$0.24$\\$\pm\,0.43$}
    & \makecell[c]{$0.76$\\$\pm\,0.43$}
    & \makecell[c]{$2710.2$\\$\pm\,2693.8$} \\
    \makecell[c]{\textbf{SAC}}
    & \makecell[c]{$0.00$\\$\pm\,0.00$}
    & \makecell[c]{$500.0$\\$\pm\,0.0$}
    & \makecell[c]{$-500.0$\\$\pm\,0.0$}
    & \makecell[c]{$0.16$\\$\pm\,0.37$}
    & \makecell[c]{$0.89$\\$\pm\,0.40$}
    & \makecell[c]{$-64.3$\\$\pm\,188.4$}
    & \makecell[c]{$0.30$\\$\pm\,0.46$}
    & \makecell[c]{$0.50$\\$\pm\,0.50$}
    & \makecell[c]{$1177.5$\\$\pm\,4484.1$} \\
    \makecell[c]{\textbf{PPO--MPPI}\\(\(\rho=0.0\))}
    & \makecell[c]{\bm{$1.00$}\\\bm{$\pm\,0.00$}}
    & \makecell[c]{\bm{$82.6$}\\\bm{$\pm\,22.9$}}
    & \makecell[c]{\bm{$-92.6$}\\\bm{$\pm\,40.1$}}
    & \makecell[c]{$0.36$\\$\pm\,0.48$}
    & \makecell[c]{$0.43$\\$\pm\,0.29$}
    & \makecell[c]{$-8.3$\\$\pm\,227.5$}
    & \makecell[c]{$0.14$\\$\pm\,0.35$}
    & \makecell[c]{$0.86$\\$\pm\,0.35$}
    & \makecell[c]{$2856.3$\\$\pm\,1465.9$} \\
    \makecell[c]{\textbf{PPO--MPPI}\\(\(\rho=0.3\))}
    & \makecell[c]{\bm{$1.00$}\\\bm{$\pm\,0.00$}}
    & \makecell[c]{$90.7$\\$\pm\,24.3$}
    & \makecell[c]{$-103.1$\\$\pm\,40.0$}
    & \makecell[c]{\bm{$0.46$}\\\bm{$\pm\,0.50$}}
    & \makecell[c]{\bm{$0.40$}\\\bm{$\pm\,0.35$}}
    & \makecell[c]{\bm{$44.2$}\\\bm{$\pm\,215.5$}}
    & \makecell[c]{\bm{$0.74$}\\\bm{$\pm\,0.44$}}
    & \makecell[c]{\bm{$0.18$}\\\bm{$\pm\,0.38$}}
    & \makecell[c]{\bm{$4512.9$}\\\bm{$\pm\,2516.0$}} \\
    \bottomrule
  \end{tabular}
  \label{tab:method_comparison_result}
\end{table}

\subsection{Comparison Result of Influence Ratio $\rho$}
\begin{figure}[t]
  \centering
  \includegraphics[width=0.95\linewidth,keepaspectratio,trim={1.8cm 0 0 0},clip]{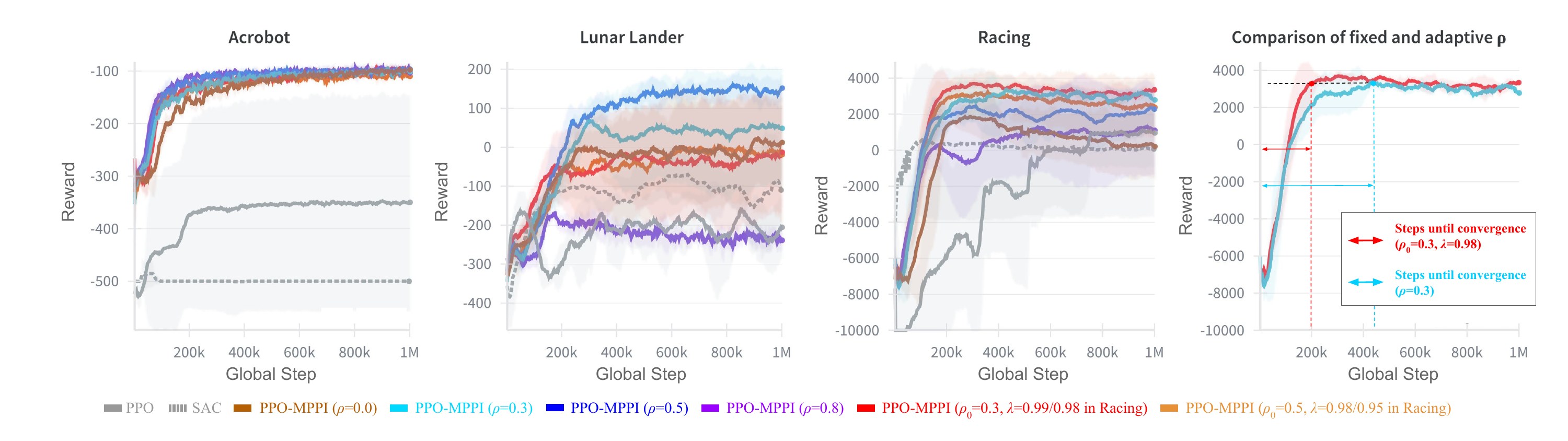}\\
  \caption{Three figures on the left: Episode reward of each environment. averaged over 5 seeds. \\
           Rightmost figure: Episode reward for PPO-MPPI($\rho=0.3$ v.s. $\rho_0=0.3, \lambda=0.98$) in Racing environment.}
  \label{fig:all_result}
\end{figure}

We next present results when the influence ratio is adapted, as detailed in Table~\ref{tab:rho_comparison_result}.
In the Acrobot domain, the final performance was nearly indistinguishable across settings (i.e., performance saturates nearly equally across all setups / parameters).  In the Lunar Lander domain, however, the best result was achieved with the fixed setting \(\rho=0.5\), with performance degrading as $\rho$ is increased to \(\rho=0.8\). 
In Racing, the best results was achieved with the \emph{adaptive influence ratio \(\rho_0=0.3,\ \lambda=0.98\)}, yielding better results than any fixed setting. This suggests that, in Racing, where model mismatch is pronounced, adapting $\rho$ mitigates model errors and improves performance.
Compared to the fixed setting (\(\rho=0.3\)), an adaptive schedule initialized at \(\rho_0=0.3\) converged faster than the remainder of the cases, reaching the highest reward achieved by the fixed setting in \(2.1\times\) fewer training steps.
In turn, when dynamics/reward models are misspecified, an environment-dependent optimal \(\rho\) exists. 
See Fig.~\ref{fig:all_result} for training curves of each of the methods.

\begin{table}[t]
  \centering
  \tiny
  \setlength{\tabcolsep}{2pt}
  \renewcommand{\arraystretch}{1.20}
  \caption{Evaluation across five influence ratios. Entries are \textit{mean} $\pm$ \textit{std} over 50 episodes per each method. $\uparrow$ / $\downarrow$ indicate whether higher / lower equates to better performance, with \textbf{bold} being best.} 
  \begin{tabular}{c c c c c c c c c c}
    \toprule
    & \multicolumn{3}{c}{\textbf{Acrobot}}
    & \multicolumn{3}{c}{\textbf{Lunar Lander}}
    & \multicolumn{3}{c}{\textbf{Racing}} \\
    \cline{2-4}\cline{5-7}\cline{8-10}
    \textbf{Method}
    & \textbf{Success Rate [$\uparrow$]} & \textbf{Step [$\downarrow$]} & \textbf{Reward[$\uparrow$]}
    & \textbf{Success Rate [$\uparrow$]} & \makecell{\textbf{Dist. to}\\\textbf{Goal [$\downarrow$]}} & \textbf{Reward[$\uparrow$]}
    & \textbf{Finish Rate [$\uparrow$]} & \makecell{\textbf{Coll./}\\\textbf{Off-track [$\downarrow$]}} & \textbf{Reward[$\uparrow$]} \\
    \midrule
    \makecell[c]{\textbf{PPO--MPPI}\\(\(\rho=0.3\))}
    & \makecell[c]{\bm{$1.00$}\\\bm{$\pm\,0.00$}}
    & \makecell[c]{$90.7$\\$\pm\,24.3$}
    & \makecell[c]{$-103.1$\\$\pm\,40.0$}
    & \makecell[c]{$0.46$\\$\pm\,0.50$}
    & \makecell[c]{$0.40$\\$\pm\,0.35$}
    & \makecell[c]{$44.2$\\$\pm\,215.5$}
    & \makecell[c]{$0.74$\\$\pm\,0.44$}
    & \makecell[c]{$0.18$\\$\pm\,0.38$}
    & \makecell[c]{$4512.9$\\$\pm\,2516.0$} \\
    \makecell[c]{\textbf{PPO--MPPI}\\(\(\rho=0.5\))}
    & \makecell[c]{\bm{$1.00$}\\\bm{$\pm\,0.00$}}
    & \makecell[c]{$88.3$\\$\pm\,27.1$}
    & \makecell[c]{$-99.3$\\$\pm\,43.5$}
    & \makecell[c]{\bm{$0.72$}\\\bm{$\pm\,0.45$}}
    & \makecell[c]{\bm{$0.15$}\\\bm{$\pm\,0.16$}}
    & \makecell[c]{\bm{$149.8$}\\\bm{$\pm\,132.3$}}
    & \makecell[c]{$0.50$\\$\pm\,0.50$}
    & \makecell[c]{$0.40$\\$\pm\,0.49$}
    & \makecell[c]{$3774.9$\\$\pm\,2669.2$} \\
    \makecell[c]{\textbf{PPO--MPPI}\\(\(\rho=0.8\))}
    & \makecell[c]{\bm{$1.00$}\\\bm{$\pm\,0.00$}}
    & \makecell[c]{\bm{$82.5$}\\\bm{$\pm\,21.7$}}
    & \makecell[c]{$-96.4$\\$\pm\,42.7$}
    & \makecell[c]{$0.06$\\$\pm\,0.24$}
    & \makecell[c]{$0.16$\\$\pm\,0.11$}
    & \makecell[c]{$-232.3$\\$\pm\,155.3$}
    & \makecell[c]{$0.44$\\$\pm\,0.50$}
    & \makecell[c]{$0.28$\\$\pm\,0.45$}
    & \makecell[c]{$2114.9$\\$\pm\,3350.9$} \\
    \makecell[c]{\textbf{PPO--MPPI}\\(\(\rho_0=0.3,\ \lambda=0.99\)\\\(\lambda=0.98\) in Racing)}
    & \makecell[c]{\bm{$1.00$}\\\bm{$\pm\,0.00$}}
    & \makecell[c]{$83.8$\\$\pm\,23.7$}
    & \makecell[c]{\bm{$-90.3$}\\\bm{$\pm\,36.0$}}
    & \makecell[c]{$0.42$\\$\pm\,0.49$}
    & \makecell[c]{$0.54$\\$\pm\,0.52$}
    & \makecell[c]{$25.9$\\$\pm\,202.6$}
    & \makecell[c]{\bm{$0.84$}\\\bm{$\pm\,0.37$}}
    & \makecell[c]{\bm{$0.16$}\\\bm{$\pm\,0.37$}}
    & \makecell[c]{\bm{$5251.5$}\\\bm{$\pm\,2045.8$}} \\
    \makecell[c]{\textbf{PPO--MPPI}\\(\(\rho_0=0.5,\ \lambda=0.98\)\\\(\lambda=0.95\) in Racing)}
    & \makecell[c]{\bm{$1.00$}\\\bm{$\pm\,0.00$}}
    & \makecell[c]{$96.6$\\$\pm\,37.6$}
    & \makecell[c]{$-115.3$\\$\pm\,62.5$}
    & \makecell[c]{$0.34$\\$\pm\,0.47$}
    & \makecell[c]{$0.48$\\$\pm\,0.30$}
    & \makecell[c]{$31.0$\\$\pm\,162.7$}
    & \makecell[c]{$0.60$\\$\pm\,0.49$}
    & \makecell[c]{$0.34$\\$\pm\,0.47$}
    & \makecell[c]{$4336.8$\\$\pm\,2130.6$} \\
    \bottomrule
  \end{tabular}
  \label{tab:rho_comparison_result}
\end{table}

\section{Conclusion}
We introduce a shared training framework that unifies the high-level reasoning capabilities of value-based RL and low-level, trajectory-based reasoning of MPPI.  We develop training scheme that leverages a bi-directional information flow between the two strategies: the RL component guides sampling, while MPPI-generated rollouts accelerate value learning and policy improvement.  We establish a bound on the value-function error that explicitly accounts for model approximation and sampling-induced bias, and propose a training algorithm featuring an adaptive influence ratio that dynamically regulates the contribution of real and virtual data based on ensemble uncertainty.  Our results across various environments underscore the role of adaptivity in mitigating model bias.

Promising future extensions could be in exploring how mixtures of real and imagined rollouts can benefit model-based or latent-world model settings (e.g., Dreamer, TD-MPC) in terms of sample efficiency and bias.  Additionally, use of uncertainty-aware constraints could benefit human-interactive and safety-critical settings more generally.

\acks{Any opinions, findings, and conclusions expressed in this material are those of the authors and do not necessarily reflect the views of TRI or any other Toyota entity.}

\bibliography{references}

\newpage

\appendix
\section{Proof of Theorem~\ref{theorem:value_error} and Additional Analyses}
\label{sec:proofs}

We note that there are three sources of bias: model mismatch between the true MDP and the mixed MDP, bias introduced by value approximation error, and a policy sub-optimality gap.



First let the compositional policy $\tilde\pi(u_t \mid s_t) := \int \pi_{\MPPI}(u_t \mid s_t ; a) \pi_\phi(a \mid s_t) da$.
We assume some environment model $\mdp$ and some virtual model used in MPPI $\hat\mdp$.  Given the interface map $\psi$ and MPPI policy $\pi_{\MPPI}(u_t \mid s_t ; a)$, with slight abuse of notation, we can recast $\hat{r}: \mathcal{S} \times \mathcal{A} \mapsto \mathbb{R}$ and 
$\hat{P}: \mathcal{S} \times \mathcal{A} \mapsto \Delta(\mathcal{S})$. 
Let $\pi: \mathcal{S} \mapsto \mathcal{A}$ be any policy.
We first introduce a Simulation Lemma to bound the error in value due to the approximations $\hat{r}$ and $\hat{P}$.

\begin{lemma}[$H$-Step Simulation Lemma]
\label{lemma:simulation}
Assume $\rho_t$ is a time-varying parameter controlling data sampled from the real environment and virtual samples $\mathcal{D}_t = (1 - \rho_t) \mathcal{D}_{\RL, t} + \rho_t \mathcal{D}_{\MPPI, t}$, 
\begin{equation}
    \begin{aligned}
        & \left\| V_{\pi, \mdp}(s) - V_{\pi, \mdp_\rho}(s) \right\|_{\infty}
        \le 
        \rho \alpha_P H \gamma \frac{1-\gamma^{H-1}}{1-\gamma}\frac{\Span(r)}{2} 
        + \rho \alpha_r \frac{1-\gamma^{H}}{1-\gamma}
    \end{aligned}
\end{equation}
\end{lemma}
\begin{proof}
Let
\begin{equation}
    \begin{aligned}
        r_\rho(s,a) &:= (1-\rho)r(s,a) + \rho\hat r(s,a), \\
        P_\rho(\cdot\mid s,a) &:= (1-\rho)P(\cdot\mid s,a) + \rho\hat P(\cdot\mid s,a)
    \end{aligned}
\end{equation}
We obtain trajectories, each containing $H$ samples.  MPC approximates the truncated-horizon value function as follows: 
\begin{equation}
    V_{\pi, P}(s) = \expectation_{\pi, P}\left[\sum_{i = 0}^{H - 1} \gamma^i r(s_i, a_i) + \gamma^H V(s_H) \middle| s_{0} = s\right]
\end{equation}

We define an MDP $\mdp_\rho = \langle\mathcal{S}, \mathcal{A}, P_\rho, r_\rho, H, \gamma, \mu\rangle$.  Assuming the actual state transitions come from the real environment $\mdp$, we can factor $r_{\rho}$, $P_{\rho}$ independently as follows:
\begin{equation}
    \begin{aligned}
        & V_{\pi, \mdp}(s) - V_{\pi, \mdp_\rho}(s) 
        = 
        V_{\pi, \mdp}(s) - \expectation_{\pi, P_\rho}\left[\sum_{i = 0}^{H - 1} \gamma^i r_\rho(s_i, a_i) + \gamma^H V(s_H)\right] \\
        & =
        V_{\pi, \mdp}(s) - \expectation_{\pi, P_\rho}\left[\sum_{i = 0}^{H - 1} \gamma^i r(s_i, a_i) + \gamma^H V(s_H)\right] \\
        & +
        \expectation_{\pi, P_\rho, \mu}\left[\sum_{i = 0}^{H - 1} \gamma^i r(s_i, a_i) + \gamma^H V(s_H)\right] - \expectation_{\pi, P_\rho, \mu}\left[\sum_{i = 0}^{H - 1} \gamma^i r_\rho(s_i, a_i) + \gamma^H V(s_H)\right]
    \end{aligned}
\end{equation}

Applying the triangle inequality yields:
\begin{equation}
    \begin{aligned}
        & \left\| V_{\pi, \mdp}(s) - V_{\pi, \mdp_\rho}(s) \right\|_{\infty}
        \\
        & \le
        \left\| \sum_{s \in \{s_0,\ldots,s_{H-1}\}} \left(P(s) - P_\rho(s)\right) \left(\sum_{i=0}^{H-1} \gamma^{i}\, r(s_i, a_i)\right) + \gamma^{H} V(s_H) \right\|_{\infty}
        \\
        & + \left\| \expectation_{\pi,P_\rho,\mu}\!\left[\sum_{i=0}^{H-1} \gamma^{i}\left(r(s_i,a_i)-r_\rho(s_i,a_i)\right)\right] \right\|_{\infty}
        \\
        & \le \rho \alpha_P H \gamma \frac{1-\gamma^{H-1}}{1-\gamma}\frac{\Span(r)}{2}
        + \rho \alpha_r \frac{1-\gamma^{H}}{1-\gamma}
    \end{aligned}
\end{equation}
Where we apply the uniform bounds $\vert r(s, a) - r_\rho(s, a)\vert \le \alpha_r$ and $\vert P(s, a) - P_\rho(s, a)\vert \le \alpha_P$, and $\Span(f) := \sup(f) - \inf(f)$.

\end{proof}

Next, we prove Theorem~\ref{theorem:value_error}.

\begin{proof}
Our goal is to bound the following triangle inequality:
\begin{equation}
    \begin{aligned}
        \left\|V_{\pi^{\star}, \mathcal{M}} - \hat{V}_{\tilde\pi, \mathcal{M}_{\rho}} \right\|_{\infty}
        \le
        \underbrace{ \left\|V_{\pi^{\star}, \mathcal{M}} - V_{\pi^{\star}, \mathcal{M}_{\rho}} \right\|_{\infty}}_{\textrm{Lemma~\ref{lemma:simulation}}}
        + 
        \underbrace{ \left\|V_{\pi^{\star}, \mathcal{M}_{\rho}} - \hat{V}_{\pi^{\star}, \mathcal{M}_{\rho}} \right\|_{\infty}}_{\textrm{Value approximation error}}
        + 
        \underbrace{ \left\|\hat{V}_{\pi^{\star}, \mathcal{M}_{\rho}} - \hat{V}_{\tilde\pi, \mathcal{M}_{\rho}} \right\|_{\infty}}_{\textrm{policy gap}}
    \end{aligned}
\end{equation}
We find bounds for the last two terms.

\paragraph{Value approximation error.} We first bound the approximation error of the terminal value function $\hat{V}$ to the true value function $V$.

Let $\varepsilon_V := \sup_s\left|V(s) - \hat{V}(s)\right|$
\begin{equation}
    \begin{aligned}
        & V_{\pi, \mdp}(s) - \hat V_{\pi, \mdp_\rho}(s) 
        = 
        V_{\pi, \mdp}(s) - \expectation_{\pi, P_\rho}\left[\sum_{i = 0}^{H - 1} \gamma^i r_\rho(s_i, a_i) + \gamma^H \hat V(s_H)\right] \\
        &= \gamma^H \left|\expectation \left[V(s_H) - \hat V(s_H) \right]\right| \le \gamma^H \varepsilon_V
    \end{aligned}
\end{equation}

\paragraph{Policy gap error.}  Our goal is to bound the error under due to the difference in policy $\tilde\pi$ from the optimal one, $\pi^{\star}$.

First, let
\begin{equation}
    \hat{V}_{\pi, \mdp}(s) = \expectation_{\pi, \mdp} \left[ \sum_{i = 0}^{H-1} \gamma^i r_{\rho}(s_i, a_i) + \gamma^H \hat{V}(s_H) \middle| s_0 = s\right]
\end{equation}
and define the advantage function as
\begin{equation}
    \hat{A}_{\pi, \mdp}(s, a) = \hat{Q}_{\pi, \mdp}(s, a) - \hat{V}_{\pi, \mdp}(s)  
\end{equation}
where
\begin{equation}
    \hat{Q}_{\pi, \mdp}(s, a) = r_{\rho}(s, a) + \gamma \expectation_{s' \sim P_{\rho}( \cdot \mid s, a)} \left[\hat{V}_{\pi, \mdp}(s') \right]
\end{equation}

Using the Performance Difference Lemma~\cite{kakade2002approximately}, we can characterize the difference in value functions of two policies $\pi_1$ and $\pi_2$ as equivalent to the sum of advantages of $\pi_2$ over time when actions are sampled from $\pi_1$:
\begin{equation}
\label{eq:val_diff_policies}
    \hat{V}_{\pi_1, \mdp_{\rho}}(s_0) - \hat{V}_{\pi_2, \mdp_{\rho}}(s_0) 
    =
    \expectation_{\pi_1, P_{\rho}} \left[\sum_{i=0}^{H-1} \gamma^i \hat{A}_{\pi_2, \mdp_{\rho}}(s_i, a_i) \middle| s_0=s\right]
\end{equation}
To find a bound for $\left\| \hat{V}_{\pi^{\star}, \mdp_{\rho}}(s) - \hat{V}_{\tilde{\pi}, \mdp_{\rho}}(s) \right\|_{\infty}$, we use the property that the advantage over $\pi$ has expectation zero when sampled over $\pi$, i.e.  $\expectation_{a\sim\pi(\cdot\mid s)}\left[\hat{A}_{\pi, \mdp_{\rho}}(s, a)\right] = 0$.  Hence,
\begin{equation}
    \left| \expectation_{a\sim\pi^{\star}(\cdot\mid s)}\left[\hat{A}_{\tilde\pi, \mdp_{\rho}}(s, a)\right]\right|
    =
    \left|\expectation_{a\sim\pi^{\star}(\cdot\mid s)}\left[\hat{A}_{\tilde\pi, \mdp_{\rho}}(s, a)\right]
    -
    \expectation_{a\sim\tilde\pi(\cdot\mid s)}\left[\hat{A}_{\tilde\pi, \mdp_{\rho}}(s, a)\right]\right|
\end{equation}
From the definition of total variation, we have, for bounded $f$:
\begin{equation}
    \left| \expectation_{a \sim\pi_1}\left[f(a)\right] - \expectation_{a \sim\pi_2}\left[f(a)\right] \right| \leq 2\left\| f(\cdot)\right\|_{\infty} \cdot D_{TV}\left(\pi_1(\cdot\mid s \Vert \pi_2(\cdot\mid s)\right)
\end{equation}
Thus, 
\begin{equation}
    \left| \expectation_{a\sim\pi^{\star}(\cdot\mid s)}\left[\hat{A}_{\tilde\pi, \mdp_{\rho}}(s, a)\right]\right|
    \leq
    2 \left\| \hat{A}_{\tilde\pi, \mdp_{\rho}}(s, a) \right\|_{\infty} \cdot D_{TV}\left(\pi_1(\cdot\mid s) \Vert \pi_2(\cdot\mid s)\right)
\end{equation}
Now, expanding the telescoping sum in \eqref{eq:val_diff_policies}, 
\begin{equation}
\label{eq:policy_value_bound}
    \begin{aligned}
        \left\| \hat{V}_{\pi^{\star}, \mdp_{\rho}} - \hat{V}_{\tilde{\pi}, \mdp_{\rho}} \right\|_{\infty}
        &= 
        \left\| \expectation_{\pi^{\star}, P_{\rho}} \left[\sum_{i=0}^{H-1} \gamma^i \hat{A}_{\tilde{\pi}, \mdp_{\rho}}(s_i, a_i) \middle| s_0=s\right] \right\|_{\infty} \\
        &\leq 
        \sum_{i=0}^{H-1} \gamma^i \left\| \expectation_{\pi^{\star}, P_{\rho}} \left[\hat{A}_{\tilde{\pi}, \mdp_{\rho}}(s_i, a_i) \middle| s_0=s\right] \right\|_{\infty} \\
        &\leq
        2 \cdot \sum_{i=0}^{H-1} \gamma^i \left\| \hat{A}_{\tilde\pi, \mdp_{\rho}}(s_i, a_i) \right\|_{\infty} \cdot D_{TV}\left(\pi^{\star}(\cdot\mid s_i) \Vert \tilde\pi(\cdot\mid s_i)\right) \\
        &\leq
        2 \cdot \frac{1 - \gamma^H}{1 - \gamma}\left\| \hat{A}_{\tilde\pi, \mdp_{\rho}}(s, a) \right\|_{\infty} \cdot \max_s D_{TV}\left(\pi^{\star}(\cdot\mid s) \Vert \tilde\pi(\cdot\mid s)\right)
    \end{aligned}
\end{equation}
To obtain a bound for $\left\| \hat{A}_{\tilde\pi, \mdp_{\rho}}(s, a) \right\|_{\infty}$, observe that
\begin{equation}
    \begin{aligned}
        \left| \hat{A}_{\tilde\pi, \mdp_{\rho}}(s, a) \right| 
        &= 
        \left| \hat{Q}_{\tilde\pi, \mdp_{\rho}}(s, a) - \hat{V}_{\tilde\pi, \mdp_{\rho}}(s) \right| 
        \leq 
        \left| \hat{Q}_{\tilde\pi, \mdp_{\rho}}(s, a) \right| + \left| \hat{V}_{\tilde\pi, \mdp_{\rho}}(s) \right| \\
    \end{aligned}
\end{equation}
Rewards are bounded by $R_{\max}$, and $\gamma < 1$, any value function satisfies $\left\|V\right\|_{\infty} \leq \frac{R_{\max}}{1 - \gamma}$. Therefore,
\begin{equation}
\left\|\hat{V}_{\tilde\pi, \mdp_{\rho}}(s)\right\|_{\infty} \leq \frac{R_{\max}}{1 - \gamma}, \qquad \left\|\hat{Q}_{\tilde\pi, \mdp_{\rho}}(s, a)\right\|_{\infty} \leq \frac{R_{\max}}{1 - \gamma}
\end{equation}
Substituting into~\eqref{eq:policy_value_bound}, we arrive at
\begin{equation}
    \left\| \hat{V}_{\pi^{\star}, \mdp_{\rho}} - \hat{V}_{\tilde{\pi}, \mdp_{\rho}} \right\|_{\infty}
    \leq
    \frac{4 R_{\max} \left(1 - \gamma^H\right)}{\left(1 - \gamma\right)^2} \cdot \max_s D_{TV}\left(\pi^{\star}(\cdot\mid s) \Vert \tilde\pi(\cdot\mid s)\right)
\end{equation}


\end{proof}
Theorem~\ref{theorem:value_error} analyzes the idealized compositional policy $\tilde{\pi}$, which assumes exact MPPI solutions. In practice, MPPI uses $K$ finite samples, and the resulting approximation error can be quantified as follows.
\begin{corollary}
\label{cor:implementation_bound}
Let $\tilde V_{\pi, \mdp_{\rho}}$ be the value that MPPI computes, and let
\begin{equation}
    g := \sum_{\tau = t}^{t + H - 1} \gamma^{\tau - t} r(s_{\tau}, a_{\tau}) + \gamma^H \hat{V}(s_{t + H}).
\end{equation}
It follows that:
\begin{equation}
    \begin{aligned}
        \left\| V_{\pi^*, \mdp}(s) - \tilde{V}_{\pi, \mdp_{\rho}}(s) \right\|_\infty
        \le
        \left\| V_{\pi^*, \mdp}(s) - \hat{V}_{\pi, \mdp_{\rho}}(s) \right\|_\infty
        +
        \Span(g) D_u \sqrt{\frac{H}{2} \lambda_{\max}(\Sigma^{-1})}
    \end{aligned}
\end{equation}
where  $\lambda_{\max}(\Sigma^{-1})$ is the largest eigenvalue of $\Sigma^{-1}$, and $D_u := \sup_{u, u' \in \mathcal{U}} \Vert u - u' \Vert$
\end{corollary}

\begin{proof}
We apply the triangle inequality.  The first term of the bound comes from Theorem~\ref{theorem:value_error}, with the second term being $\left\| \hat{V}_{\pi, \mdp_{\rho}}(s) - \tilde{V}_{\pi, \mdp_{\rho}}(s) \right\|_\infty$.  To show the second term, we form a distributional / rescoring bound on the samples drawn by MPPI.
Given two distributions $q$ and $q'$ and a function $g$, we can construct a bound
\begin{equation}
\label{eq:pinsker}
    \expectation_{q}[g] - \expectation_{q'}[g] \le \Span(g)\sup_{s \in \mathcal{S}} \left|q(s) - q'(s)\right| \le \Span(g) \sqrt{\frac{1}{2} D_{KL}(q \Vert q')}
\end{equation}
where the inequalities, respectively, follow from total variation and Pinsker's inequality.


Under the sampling rule in~\eqref{eq:mppi_update_law}, a nominal control sequence $\bar{u}_{t:t+H-1}$ is perturbed by noise of the form $\varepsilon_{\tau}^i \sim \mathcal{N}(0, \Sigma)$. A weighted average of these perturbations are constructed at time $t$ as:
\begin{equation}
    u_t^{(m)} = \bar u_t + \sum_{k=1}^K \tilde{w}_k^{(m)} \varepsilon_t^{(k)}
\end{equation}
For the chosen candidate, let the difference of candidate $m$'s sequence from the nominal sequence be defined as $\Delta u_t^{(m)} := u_t^{(m)} - \bar u_t$.  Let
\begin{equation}
    g_{\pi}(s_0) := \sum_{i = 0}^{H - 1} \gamma^i r(s_i, a_i) + \gamma^H \hat{V}(s_H)
\end{equation}
Define two distributions $u_t \sim q_0$ and $u_t^{(m)} \sim q_m$, respectively, as the nominal distribution for $u_t$ and the MPPI-sampled control $u_t^{(m)}$. Then,~\eqref{eq:pinsker} gives:
\begin{equation}
    \begin{aligned}
        &\expectation_{q_m}[g_{\pi}(s_0)] - \expectation_{q_0}[g_{\pi}(s_0)] \le \Span(g_{\pi}(s_0))\sqrt{\frac{1}{2} D_{KL}(q_m \Vert q_0)} \\
        &= \Span(g_{\pi}(s_0)) \sqrt{\frac{1}{2} D_{KL}\left(\mathcal{N}\big(\Delta u_{\tau}^{(m)}, \Sigma\big) \Vert \mathcal{N}(0, \Sigma)\right)} \\
        &= \Span(g_{\pi}(s_0)) \sqrt{\frac{1}{2} \sum_{\tau=t}^{t + H - 1} \left(\Delta u_{\tau}^{(m)}\right)^T \Sigma^{-1} \Delta u_t^{(m)}}
    \end{aligned}
\end{equation}
Note that the above bound is empirical, requiring a set of MPPI samples.  To find a non-empirical bound, we can observe the following property:
\begin{equation}
    \left(\Delta u_{\tau}^{(m)}\right)^T \Sigma^{-1} \Delta u_t^{(m)} 
    \le \lambda_{\max}(\Sigma^{-1}) \Vert \Delta u_{\tau}^{(m)} \Vert^2 
    \le \lambda_{\max}(\Sigma^{-1}) D_u^2
\end{equation}
where $\lambda_{\max}(\cdot)$ is the largest eigenvalue of a matrix, and diameter $D_u := \sup_{u, u' \in \mathcal{U}} \Vert u - u' \Vert$.  Hence,
\begin{equation}
    \sum_{\tau=t}^{t + H - 1} \left(\Delta u_{\tau}^{(m)}\right)^T \Sigma^{-1} \Delta u_t^{(m)}
    \le H \lambda_{\max}(\Sigma^{-1}) D_u^2
\end{equation}


\end{proof}

It is worth noting that the bound within Corollary~\ref{cor:implementation_bound} does not decay as modeling error improves.  However, the corollary makes it clear that increasing $K$ (more MPPI samples) or decreasing $\Sigma$ tightens this term, giving concrete knobs to reduce the gap between the idealized and implemented algorithms.

\section{Domain Details}
For each domain, we introduce relevant details for the environment and provide further details about approximations and choices used for MPPI.
\subsection{Acrobot}
\paragraph{Task.}
Swing up the end effector of a two-link underactuated pendulum to a target height by applying a continuous torque at the actuated joint, while avoiding a pre-specifed set of danger zones.

\paragraph{Observation space.}
A 6-dimensional continuous vector
\([ \cos\theta_1,\ \sin\theta_1,\ \cos\theta_2,\ \sin\theta_2,\ \dot\theta_1,\ \dot\theta_2 ]\),
where \(\theta_1\) is the absolute angle of the first link and \(\theta_2\) is the angle of the second link relative to the first.
In addition, the observation includes the \emph{DangerZone} parameters \((x_D, y_D, s_D)\), which specify the center coordinates and the side length of a square danger region (in the same coordinate frame as \(x,y\)).

\paragraph{Action space.}
A scalar continuous torque command at the actuated joint,
\(u \in [-1,1]\ \mathrm{N\,m}\).

\paragraph{Reward.}
\begin{itemize}
\item \(-1\) per time step until termination (shorter episodes yield higher return).
\item Additional \(-50\) per time step while the agent is inside the \emph{DangerZone}.
\end{itemize}

\paragraph{Termination.}
\begin{itemize}
\item Goal height achieved: \(-\cos\theta_1-\cos(\theta_1+\theta_2) > 1.0\).
\item Otherwise, the episode is truncated at a fixed horizon (e.g., 500 steps).
\end{itemize}

\paragraph{Dynamics Model and reward model for MPPI.}
\begin{itemize}
\item Dynamics model: exactly same as the environment's dynamics.
\item Reward model: exactly same as the environment's reward structure. 
\end{itemize}

\subsection{LunarLander}
\paragraph{Task.}
Control a lunar lander to achieve a soft touchdown at the center of a landing pad while minimizing fuel use and impact velocity, while avoiding a randomly-selected set of danger zones.

\paragraph{Observation space.}
The observation consists of the 8-dimensional state \\
\([x,\ y,\ \dot x,\ \dot y,\ \phi,\ \dot\phi,\ \texttt{leg\_left},\ \texttt{leg\_right}]\),
where the last two entries are binary flags indicating ground contact for the left and right legs,
augmented with the \emph{DangerZone} parameters \((x_D, y_D, w_D, h_D)\) that give the center coordinates and the width/height of a rectangular danger region (in the same coordinate frame as \(x,y\)).  Note that the \emph{DangerZones} in this domain are randomized.

\paragraph{Action space.}
A 2-D continuous control vector \(u=[u_{\text{main}},\,u_{\text{lateral}}] \in [-1,1]^2\).
The first element controls the main engine throttle (values \(\le 0\) are treated as off), and the second controls the lateral thrusters; its sign indicates left (\(<0\)) or right (\(>0\)) firing.

\paragraph{Reward.}
\begin{itemize}
\item Increased as the lander gets closer to the center of the landing pad, and decreased as it moves farther away.
\item Increased as the lander’s translational velocity becomes slower, and decreased as it moves faster.
\item Decreased the more the lander is tilted (i.e., larger absolute tilt angle \(|\phi|\)).
\item Increased by \(+10\) points for each leg in ground contact.
\item Decreased by \(0.03\) points per time step while any side engine is firing.
\item Decreased by \(0.3\) points per time step while the main engine is firing.
\item Additional \(-100\) for crashing and \(+100\) for a safe landing (episode end).
\item Additional \(-5\) per time step while the agent is inside the \emph{DangerZone}.
\end{itemize}

\paragraph{Termination.}
\begin{itemize}
\item Crash (lander body contacts the surface).
\item Leaving the viewport bounds.
\item Lander is not awake (comes to rest).
\item Otherwise, the episode is truncated at a fixed horizon (maximum number of steps).
\end{itemize}

\paragraph{Dynamics Model and reward model for MPPI.}
\begin{itemize}
\item Dynamics model: We omit Box2D’s complex rigid-body calculations and instead update the vehicle’s position and orientation using fixed constants.
\item Reward model: We exclude the terminal reward from the one in the environment.  Terminal costs in this domain tend to outweighs the dense reward signals.
\end{itemize}

\subsection{Racing}
\paragraph{Task.}
Drive along a designated track to reach the goal as quickly as possible while avoiding boundary violations and collisions with other vehicles, and avoid pre-specified danger zones.

\paragraph{Observation space.}
A concatenation of the following components:
\begin{itemize}
  \item Vehicle state: planar position \(x,y\), speed \(v\) (and optionally heading \(\psi\), yaw rate \(r\), etc.).
  \item TrackState: boundary deviation coefficient \(b\) (normalized lateral offset from the track centerline) and heading error \(\Delta\psi\) between the vehicle heading and the track tangent.
  \item TrackPoint: local boundary information (e.g., a fixed-length set of nearby boundary points or left/right distances) expressed in the vehicle frame.
  \item DangerZone: center coordinates \((x_D, y_D)\) and side length \(s_D\) of a square danger region.
\end{itemize}

\paragraph{Action space.}
Steering and throttle commands
\(u = [u_{\text{steer}},\,u_{\text{throt}}] \in [-1,1]^2\).

\paragraph{Reward.}
\begin{itemize}
  \item Passing reward: increased with the vehicle’s relative speed to a nearby \emph{OtherVehicle} when overtaking (i.e., larger forward relative velocity yields larger reward).
  \item Progress reward: proportional to the forward arc-length progress along the track centerline.
  \item Boundary penalty: decreased according to the magnitude of the boundary deviation coefficient \(b\).
  \item Additional \(-100\) upon exceeding boundary limits (hard off-track violation).
  \item Additional \(-1000\) upon collision with another vehicle.
  \item Additional \(+1000\) upon reaching the goal.
  \item Additional \(-150\) per time step while the agent is inside the \emph{DangerZone}.
\end{itemize}

\paragraph{Termination.}
\begin{itemize}
  \item Exceeding boundary limits (off-track).
  \item Collision with another vehicle.
  \item Goal reached.
  \item Otherwise, the episode is truncated at a fixed time limit (maximum number of steps).
\end{itemize}

\paragraph{Dynamics Model and reward model for MPPI.}
\begin{itemize}
\item Dynamics model: We apply the kinematic bicycle model (\cite{polack2017kinematic}).
\item Reward model: We exclude the terminal reward from the one in the environment.
\end{itemize}

\section{Architecture Details}
We adopt Stable-Baselines3 (SB3)~\cite{stable-baselines3} for our reinforcement learning implementations; the hyperparameters are listed in Tables~\ref{tab:ppo_params} and~\ref{tab:sac_params}.
For MPPI, we use \href{https://pypi.org/project/pytorch-mppi/0.4.1/}{pytorch-mppi}; the hyperparameters are listed in Table~\ref{tab:mppi_params}.
The MPPI running cost is decomposed as
\begin{equation}
  J_i \;=\; \sum_{\tau=t}^{t+H-1}
  \Big(
      w_{\mathrm{RL}}\, J_{\mathrm{RL}}(\hat x_\tau^{i},u_\tau^{i};\,a_t)
      \;+\;
      w_{\mathrm{D}}\, J_{\mathrm{danger}}(\hat x_\tau^{i})
      \;+\;
      J_{\mathrm{other}}(\hat x_\tau^{i},u_\tau^{i})
  \Big),
  \label{eq:mppi_cost_decomp}
\end{equation}
where \(w_{\mathrm{RL}}, w_{\mathrm{D}} \in \mathbb{R}_{\ge 0}\) are fixed weights.

\paragraph{RL term.}
Three common forms are used for the RL term. (i) is used for the results in Section~\ref{sec:result}, and (ii), (iii) are used for the results in Section~\ref{subsec:additional_result}. \\
\textit{(i) Target-tracking form:}
\begin{equation}
  J_{\mathrm{RL}}^{\mathrm{track}}\!(\hat x_\tau^{i};\,a_t)
  \;=\;
  \big\|\hat x_{\tau}^{i} - x^{\ast}(a_t)\big\|
\end{equation}
In Acrobot, $x^{\ast}(a_t) = [\theta_1^{\ast}, \theta_2^{\ast}]$, and in Lunar Lander and Racing, $x^{\ast}(a_t) = [\dot x^{\ast}, \dot y^{\ast}]$.\\
\textit{(ii) Quadratic (QP-like) form:}
\begin{equation}
\label{eq:qp_cost}
  J_{\mathrm{RL}}^{\mathrm{quad}}\!(\hat x_\tau^{i},u_\tau^{i};\,a_t)
  \;=\;
  \begin{bmatrix}\hat x_\tau^{i}\\[2pt] u_\tau^{i}\end{bmatrix}^{\!\top}
  Q_{\tau}(a_t)
  \begin{bmatrix}\hat x_\tau^{i}\\[2pt] u_\tau^{i}\end{bmatrix}
  \;+\;
  p_{\tau}^{\top}(a_t)
  \begin{bmatrix}\hat x_\tau^{i}\\[2pt] u_\tau^{i}\end{bmatrix},
\end{equation}
where $(Q_\tau,p_\tau)$ or the reference $x^\ast(a_t)$ can be viewed as quantities parameterized by the RL output $a_t$.\\
\textit{(iii) Value function terminal cost form:}
\begin{equation}
\label{eq:v_cost}
  J_{\mathrm{RL}}^{\mathrm{V}}\!(\hat x_H^{i})  \;=\;  -V(\hat x_H^{i})
\end{equation}
Note that we can combine (iii) with either (i) or (ii).

\paragraph{Danger-zone term.}
Let the axis-aligned danger zone be the rectangle centered at $(x_D,y_D)$ with width $W$ and height $H$:
\(
\mathcal{D} = \{(x,y): |x-x_D|\le W/2,\ |y-y_D|\le H/2\}
\)
(the square case is \(W=H\)).
We use a binary indicator:
\begin{equation}
  J_{\mathrm{danger}}(\hat x_\tau^{i})
  \;=\;
  \begin{cases}
    1, & (x(\hat x_\tau^{i}),y(\hat x_\tau^{i})) \in \mathcal{D},\\
    0, & \text{otherwise.}
  \end{cases}
\end{equation}

\paragraph{Other term.}
Below we specify the instances used in each domain.

\medskip
\noindent\emph{Lunar Lander.}
We penalize altitude from the pad ($y=0$ at the pad) and control effort:
\begin{equation}
  J_{\mathrm{other}}^{\mathrm{LL}}(\hat x_\tau^{i},u_\tau^{i})
  \;=\;
  w_{y}\, \bigl(y(\hat x_\tau^{i})\bigr)^{2}
  \;+\; w_{\mathrm{act}}\, \|u_\tau^{i}\|_{2}^{2}.
\end{equation}

\medskip
\noindent\emph{Racing.}
We combine a collision indicator with a boundary-violation penalty.
Let $p(\hat x)=(x(\hat x),y(\hat x))$ be the ego position, $p_{\tau,\mathrm{opponent}}^i$ the opponent position, and $d_{\mathrm{contact}}$ the contact threshold (accounting for vehicle sizes). Define the collision cost
\begin{equation}
  J_{\mathrm{coll}}(\hat s_\tau^{i})
  \;=\;
  \begin{cases}
    1, & \bigl\|p(\hat x_\tau^{i})-p_{\tau,\mathrm{opponent}}^i\bigr\|_{2} \le d_{\mathrm{contact}}\\
    0, & \text{otherwise.}
  \end{cases}
\end{equation}
and the boundary proportion $\mathrm{BP}(\hat x)$ as the radial distance from the lane center normalized by the center-to-edge distance (so $\mathrm{BP}=1$ at the boundary). The boundary penalty is
\begin{equation}
  J_{\mathrm{bound}}(\hat x_\tau^{i})
  \;=\;
  \bigl[\max\!\bigl(0,\, \mathrm{BP}(\hat x_\tau^{i}) - 1\bigr)\bigr]^{2}.
\end{equation}
We set
\begin{equation}
  J_{\mathrm{other}}^{\mathrm{Race}}(\hat x_\tau^{i})
  \;=\;
  w_{\mathrm{coll}}\, J_{\mathrm{coll}}(\hat x_\tau^{i})
  \;+\;
  w_{\mathrm{bound}}\, J_{\mathrm{bound}}(\hat x_\tau^{i}).
\end{equation}
where $w_y, w_{\mathrm{act}}, w_{\mathrm{coll}}, w_{\mathrm{bound}} \in \mathbb{R}_{\ge 0}$ are fixed scalar weights.

\begin{table}[t]
    \begin{minipage}{0.5\textwidth}
        \centering
        \caption{PPO Hyperparameters}
        \begin{tabular}{ll}
            \toprule
            \textbf{Parameter} & \textbf{Default} \\
            \midrule
            learning\_rate & $3\times10^{-4}$ \\
            n\_steps & 2048 \\
            batch\_size & 64 \\
            n\_epochs & 10 \\
            gamma & 0.99 \\
            gae\_lambda & 0.95 \\
            clip\_range & 0.2 \\
            ent\_coef & 0.0 \\
            vf\_coef & 0.5 \\
            \bottomrule
        \end{tabular}
        \label{tab:ppo_params}
    \end{minipage}
    \begin{minipage}{0.5\textwidth}
        \centering
        \caption{SAC Hyperparameters}
        \begin{tabular}{ll}
            \toprule
            \textbf{Parameter} & \textbf{Default} \\
            \midrule
            learning\_rate & $3\times10^{-4}$ \\
            buffer\_size & 1000000 \\
            batch\_size & 256 \\
            tau & 0.005 \\
            gamma & 0.99 \\
            train\_freq & 1 \\
            gradient\_steps & 1 \\
            ent\_coef & ``auto'' \\
            target\_update\_interval & 1 \\
            \bottomrule
        \end{tabular}
        \label{tab:sac_params}
    \end{minipage}
\end{table}

\begin{table}[t]
    \centering
    \caption{MPPI Hyperparameters}
    \label{tab:mppi_params}
    \begin{tabular}{llccc}
        \toprule
        \textbf{Parameter} & \textbf{Acrobot} & \textbf{LunarLander} & \textbf{Racing} \\
        \midrule
        horizon (state ref) & 10 & 10 & 10 \\
        horizon (QP) & 3   & 3   & 2   \\
        noise\_sigma & 0.5 & 0.5 & 0.5 \\
        num\_samples & 100 & 100 & 100 \\
        lambda       & 1.0 & 1.0 & 1.0 \\
        \midrule
        $w_{\mathrm{RL}}$     & 50.0        & 50.0        & 1.0 \\
        $w_{\mathrm{D}}$      & 50.0        & 400.0       & 300.0 \\
        $w_{\mathrm{act}}$    & \textemdash & 20.0        & \textemdash \\
        $w_{\mathrm{y}}$      & \textemdash & 10.0        & \textemdash \\
        $w_{\mathrm{bound}}$  & \textemdash & \textemdash & 500.0 \\
        $w_{\mathrm{coll}}$   & \textemdash & \textemdash & 300.0 \\
        \bottomrule
    \end{tabular}
    \vspace{0.25em}
\end{table}

\section{Additional Results}
\label{subsec:additional_result}
We first evaluate alternative cost formulations to show that our methods are not tied to a particular cost form.  We further show results using per-sample mixing, as described by~\eqref{eq:sample_mixing}, to contrast with the loss function weighting approach described by~\eqref{eq:loss_weighting}.

\begin{figure}[t]
  \centering
  \includegraphics[width=\linewidth]{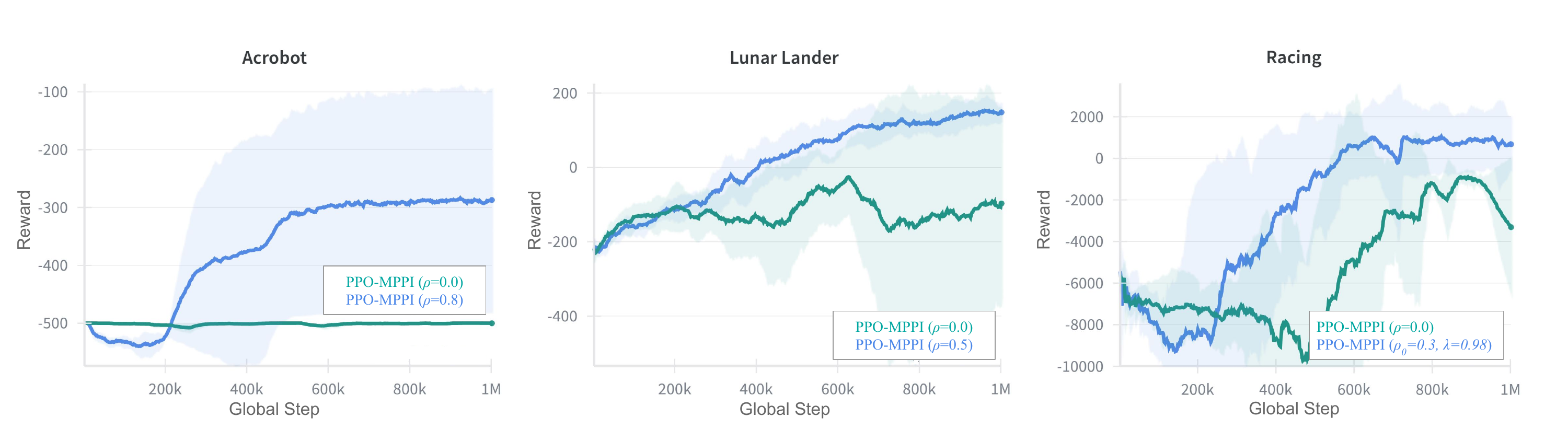}
  \caption{Episode reward under the quadratic (QP) cost formulation averaged over 5 seeds.}
  \label{fig:add_result_QP}
\end{figure}

\begin{figure}[t]
  \centering
  \includegraphics[width=\linewidth]{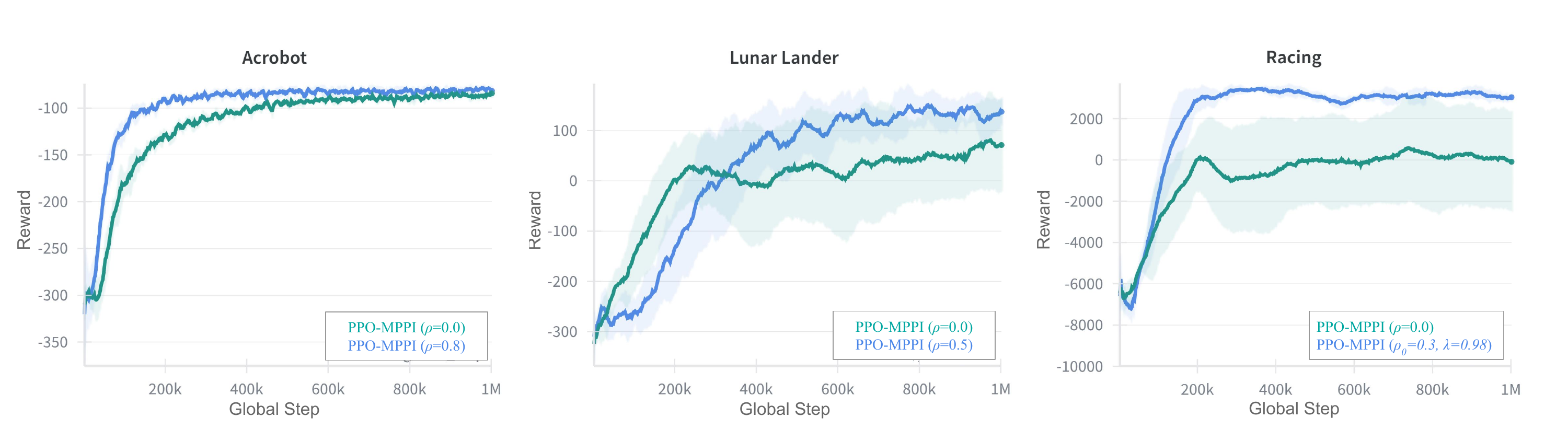}
  \caption{Episode reward under using the RL value function \(V\) as the terminal cost averaged over 5 seeds.}
  \label{fig:add_result_V}
\end{figure}

\paragraph{Alternative cost formulations} Fig.~\ref{fig:add_result_QP}, \ref{fig:add_result_V} report two evaluations, each comparing training \emph{with} vs.\ \emph{without} MPPI virtual rollout data.
Fig.~\ref{fig:add_result_QP} uses the quadratic cost formulation (\ref{eq:qp_cost}), and Fig.~\ref{fig:add_result_V} uses the RL value function as a terminal cost in MPPI (\ref{eq:v_cost}), a common way to couple RL and MPPI.
Across both formulations and all environments, we observe that adding MPPI virtual rollout data consistently yields higher performance and better sample efficiency.

\begin{figure}[t]
  \centering
  \includegraphics[width=0.95\linewidth,keepaspectratio,trim={1.8cm 0 0 0},clip]{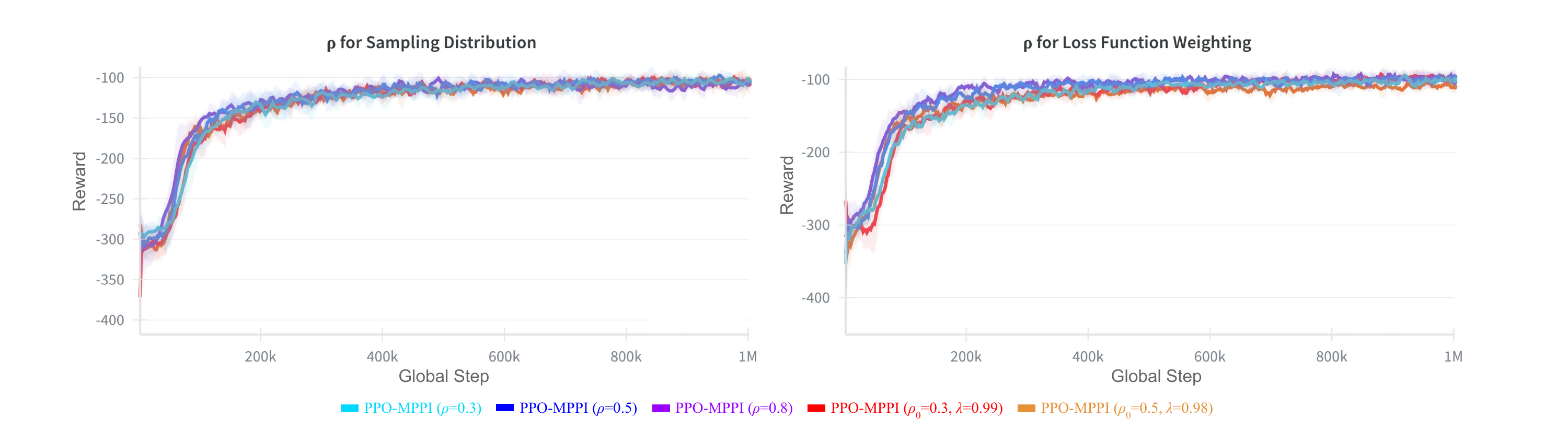}\\
  \caption{Comparison of applying the influence ratio $\rho$ to the sampling distribution (\ref{eq:sample_mixing}) v.s. the loss weighting (\ref{eq:loss_weighting}), averaged over 5 seeds in Acrobot environment.}
  \label{fig:dist_los_compare_acrobot}
\end{figure}

\begin{figure}[t]
  \centering
  \includegraphics[width=0.95\linewidth,keepaspectratio,trim={1.8cm 0 0 0},clip]{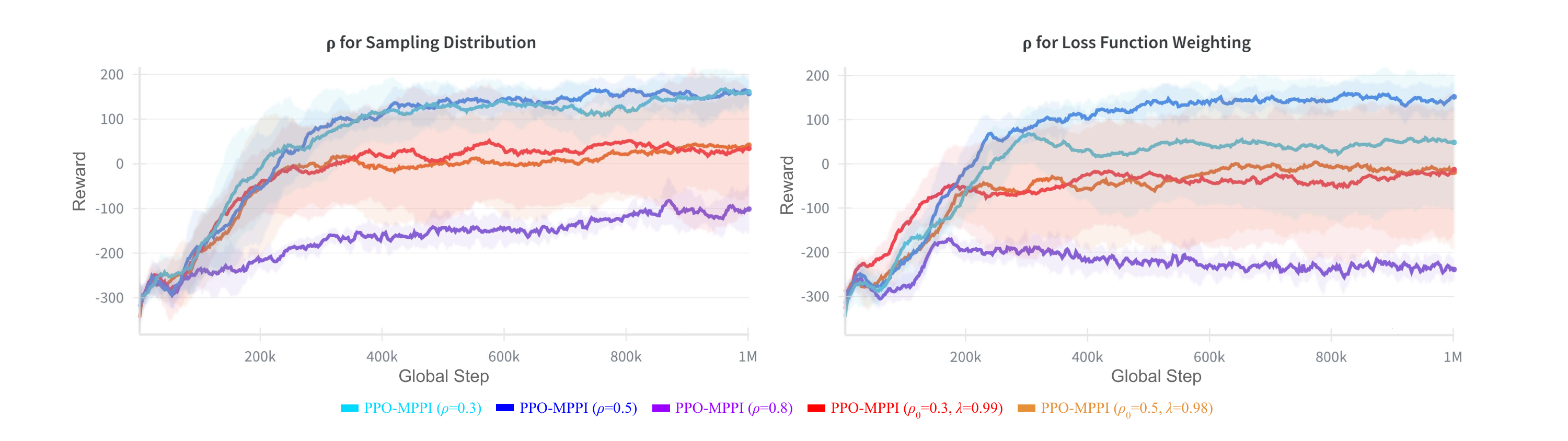}\\
  \caption{Comparison of applying the influence ratio $\rho$ to the sampling distribution (\ref{eq:sample_mixing}) vs. the loss weighting (\ref{eq:loss_weighting}), averaged over 5 seeds in Lunar Lander environment.}
  \label{fig:dist_los_compare_lunarlander}
\end{figure}

\begin{figure}[t]
  \centering
  \includegraphics[width=0.95\linewidth,keepaspectratio,trim={1.8cm 0 0 0},clip]{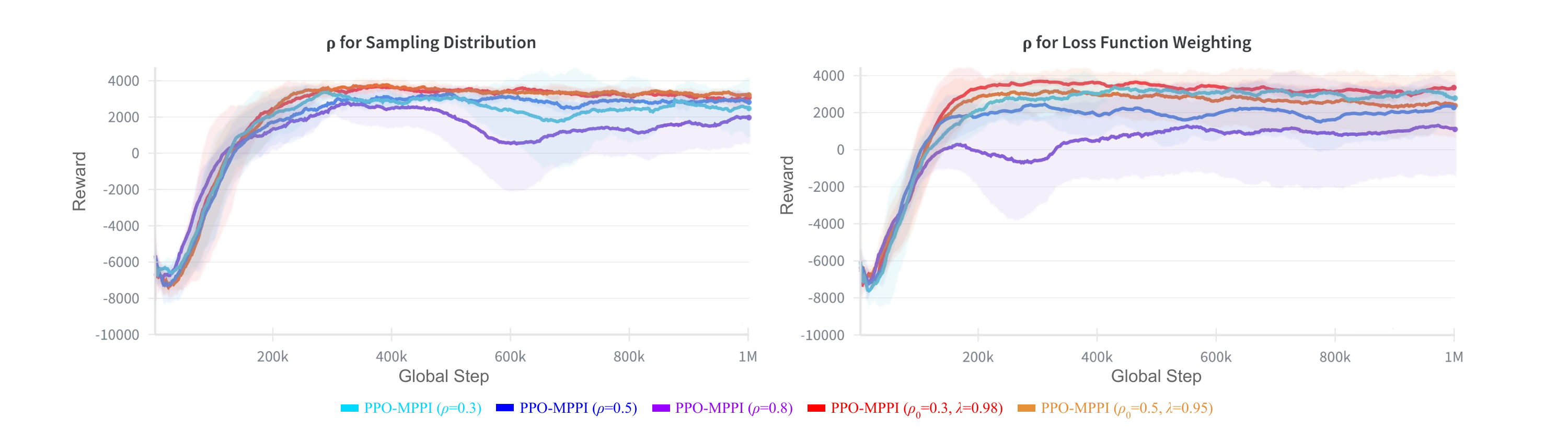}\\
  \caption{Comparison of applying the influence ratio $\rho$ to the sampling distribution (\ref{eq:sample_mixing}) vs. the loss weighting (\ref{eq:loss_weighting}), averaged over 5 seeds in Racing environment.}
  \label{fig:dist_los_compare_racing}
\end{figure}

\paragraph{Per-sample mixing} We also show in Fig.~\ref{fig:dist_los_compare_acrobot}, ~\ref{fig:dist_los_compare_lunarlander}, ~\ref{fig:dist_los_compare_racing} that applying the influence ratio $\rho$ to the sampling distribution (\ref{eq:sample_mixing}) yields the similar qualitative trend as shown in the main result in Fig.~\ref{fig:all_result} (reproduced here for comparison), which uses loss weighting (\ref{eq:loss_weighting}). 
Table~\ref{tab:welch-t-test-all} summarizes the results of Welch's $t$-tests, reporting the $t$-statistics and $p$-values when comparing the baseline setting of $\rho$ against other configurations for both $\rho$ applied to sampling distribution of~\eqref{eq:sample_mixing} and $\rho$ applied to loss~\eqref{eq:loss_weighting} in each environment.  In the Acrobot environment, most of the $p$-values are sufficiently large, and in particular no significant differences are observed for any setting of $\rho$ for dist. This indicates that performance does not change much with different choices of $\rho$, and this trend itself is shared by both $\rho$ for dist and $\rho$ for loss. In contrast, in the LunarLander environment, almost all settings exhibit significant differences for both $\rho$ for dist and $\rho$ for loss, and the signs of the $t$-values coincide across settings. Therefore, the ranking of which $\rho$ values perform better is very similar between the two methods. In the Racing environment, the presence or absence of significance and the signs of the $t$-values agree between the two metrics for three of the four settings, so the overall trend is broadly similar; however, for the setting $\rho_0 = 0.5, \lambda = 0.95$ the signs of the $t$-values for $\rho$ for sampling distribution and $\rho$ for loss are reversed, indicating a clear discrepancy in this case. This discrepancy is likely due to differences in both the amount and the bias of the data used for policy updates between $\rho$ for distribution and $\rho$ for loss. Concretely, with $\rho$ for distribution, at each step samples are stored only in one of $D_{\mathrm{rl}}$ or $D_{\mathrm{mppi}}$, whereas with $\rho$ for loss, at each step as many samples as the number of RL outputs are stored, with one sample saved in $D_{\mathrm{rl}}$ and all remaining samples saved in $D_{\mathrm{mppi}}$. As a result, the total amount of data used for $\rho$ for loss becomes larger. In a simple environment such as Acrobot, the state distribution has small variation and the data bias is limited, so this difference in data volume has only a minor impact on performance, which appears as a common trend that \textit{changing $\rho$ does not substantially affect performance.} In contrast, in a complex environment such as Racing, the state space is larger and the collected data is more prone to bias, so the amount of data accumulated in each buffer has a strong effect on performance. This can explain why, as observed, the trends for $\rho$ for distribution and $\rho$ for loss diverge for some settings. LunarLander can be seen as an environment of intermediate difficulty, where both metrics obtain sufficient data without severe bias, leading to the very similar trends observed between the two methods.

\begin{table}[t]
  \centering
  \small
  \setlength{\tabcolsep}{2pt}
  \renewcommand{\arraystretch}{1.20}
  \caption{Welch t test for all tasks.}
  \begin{tabular}{c c l c c c c}
    \toprule
    & & &
    \multicolumn{2}{c}{\textbf{$\rho$ for dist}} &
    \multicolumn{2}{c}{\textbf{$\rho$ for loss}} \\
    \cline{4-5}\cline{6-7}
    \textbf{Env} & \textbf{Baseline} & \textbf{Setting}
    & \textbf{$t$} & \textbf{$p$}
    & \textbf{$t$} & \textbf{$p$} \\
    \midrule
    \multirow{4}{*}{Acrobot}
      & \multirow{4}{*}{$\rho_0=0.3,\ \lambda=0.99$}
      & $\rho=0.3$
      & 0.643  & 0.522  & 3.31  & 0.133e-2  \\
    & & $\rho=0.5$
      & 0.0318 & 0.975  & 0.481 & 0.632     \\
    & & $\rho=0.8$
      & 1.56   & 0.123  & 1.11  & 0.271     \\
    & & $\rho_0=0.5,\ \lambda=0.98$
      & 0.156  & 0.876  & 7.54  & 0.276e-10 \\
    \midrule
    \multirow{4}{*}{Lunar-lander}
      & \multirow{4}{*}{$\rho = 0.5$}
      & $\rho=0.3$
      & 0.423 & 0.673     & 4.74  & 0.162e-4  \\
    & & $\rho=0.8$
      & 35.3  & 0.469e-53 & 67.8  & 0.511e-81 \\
    & & $\rho_0=0.3,\ \lambda=0.99$
      & 6.83  & 0.737e-8  & 8.65  & 0.858e-11 \\
    & & $\rho_0=0.5,\ \lambda=0.98$
      & 8.00  & 0.646e-10 & 7.18  & 0.233e-8  \\
    \midrule
    \multirow{4}{*}{Racing}
      & \multirow{4}{*}{$\rho_0=0.3,\ \lambda=0.98$}
      & $\rho=0.3$
      & 2.44  & 0.179e-1 & 15.1 & 0.108e-24 \\
    & & $\rho=0.5$
      & 2.99  & 0.374e-2 & 7.22 & 0.263e-8  \\
    & & $\rho=0.8$
      & 5.59  & 0.790e-6 & 6.64 & 0.235e-7  \\
    & & $\rho_0=0.5,\ \lambda=0.95$
      & -4.47 & 0.299e-4 & 3.70 & 0.541e-3  \\
    \bottomrule
  \end{tabular}
  \label{tab:welch-t-test-all}
\end{table}

\section{Additional MPPI Results}
In this section, we provide additional analyses of the MPPI component used in the proposed method using Lunar Lander environment. Specifically, we investigate (1) the increase in computational cost introduced by MPPI, (2) the sensitivity of performance to key MPPI hyperparameters, and (3) the performance of MPPI when used alone.

\paragraph{Computational cost.}
We first evaluate the additional computational cost introduced by adding MPPI to the PPO framework. Table~\ref{tab:calc_time} compares the average computation time of the pure RL baseline and PPO-MPPI for 1M training steps used in the main results. As expected, incorporating MPPI increases the computational burden because each control step requires sampling-based trajectory optimization.

This result indicates that the performance gain of PPO-MPPI comes at the cost of additional computation. However, as shown in the main results, PPO-MPPI also provides improved control performance, suggesting that this increased cost should be interpreted as a trade-off for better decision quality. Importantly, in our setting, the increase in computation remained within a controllable range and scaled predictably with MPPI hyperparameters such as the prediction horizon and the number of samples.

\begin{table}[t]
    \centering
    \caption{Comparison of computation time between PPO and PPO-MPPI.}
    \label{tab:calc_time}
    \begin{tabular}{lcc}
        \toprule
        \textbf{Domain} & \textbf{PPO} & \textbf{PPO-MPPI} \\
        \midrule
        Acrobot       & 33m21s   & 15h9m58s   \\
        Lunar Lander  & 48m45s   & 5h53m20s   \\
        Racing        & 5h34m31s & 2d2h7m45s  \\
        \bottomrule
    \end{tabular}
    \vspace{0.25em}
\end{table}

\begin{figure}[t]
  \centering
  \begin{minipage}[t]{0.48\textwidth}
    \centering
    \includegraphics[height=0.17\textheight]{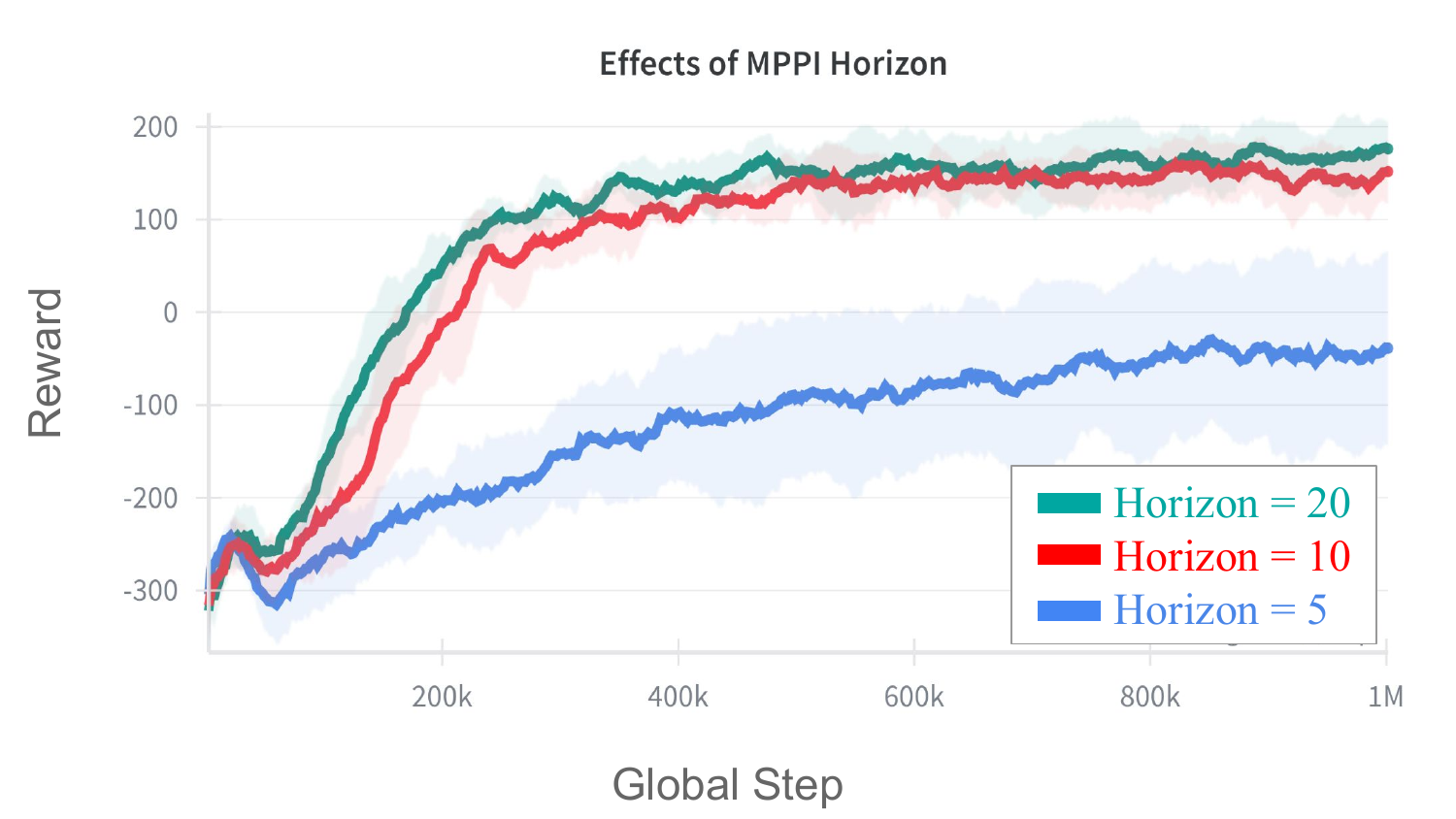}\\
    \small (a) Effects of MPPI horizon.
  \end{minipage}\hfill
  \begin{minipage}[t]{0.48\textwidth}
    \centering
    \includegraphics[height=0.17\textheight]{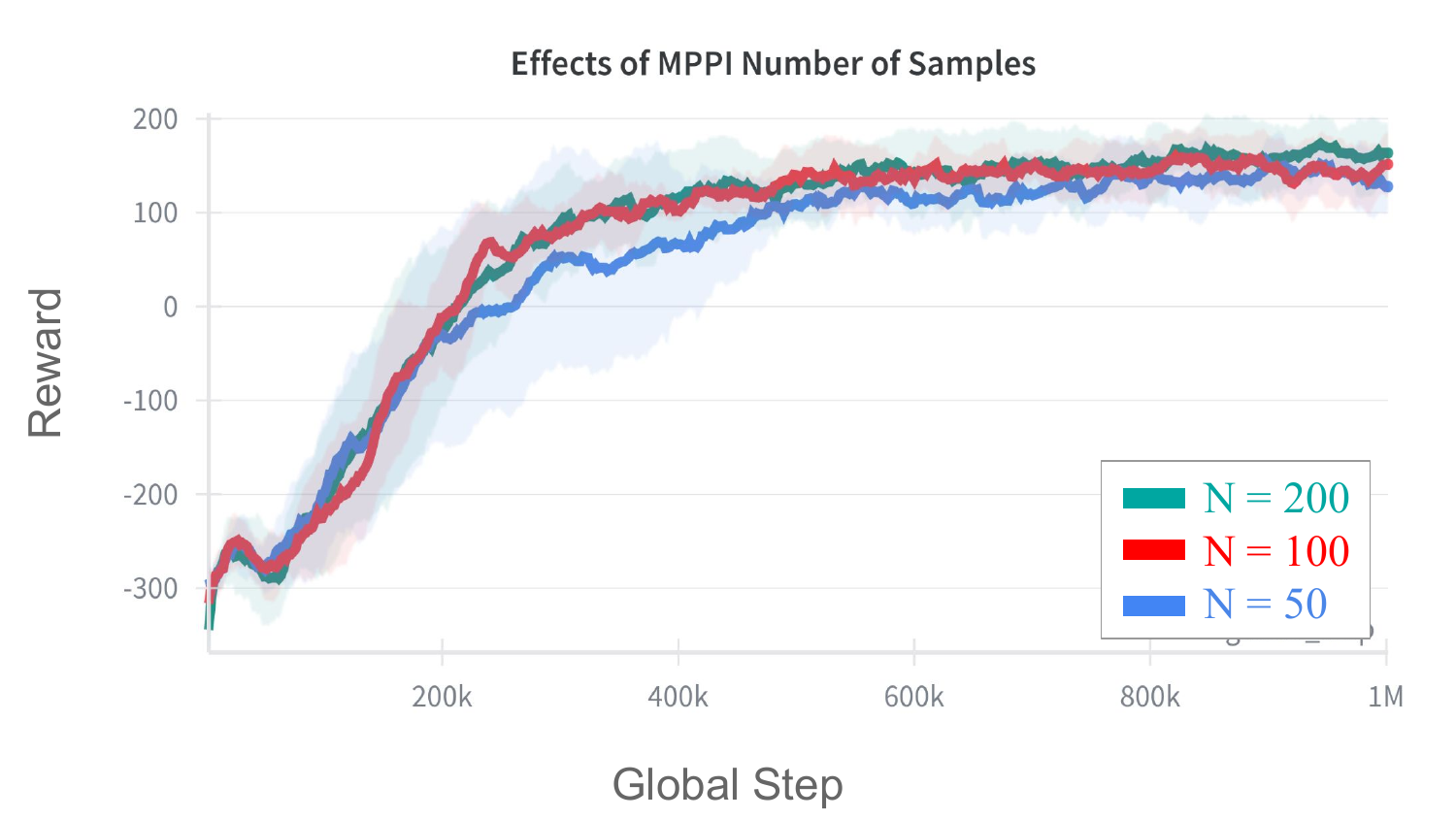}\\
    \small (b) Effects of the number of MPPI samples.
  \end{minipage}
  \caption{Sensitivity analysis of key MPPI hyperparameters.}
  \label{fig:mppi_hyperparameters}
\end{figure}

\paragraph{Sensitivity to MPPI hyperparameters.}
Next, we investigate the sensitivity of the proposed framework to two key MPPI hyperparameters: the prediction horizon and the number of samples. Fig.~\ref{fig:mppi_hyperparameters} overlays the learning curves obtained under different parameter settings with those of the default configuration shown in Table~\ref{tab:mppi_params}.

\textit{Prediction horizon.}
We first examine the effect of the prediction horizon. When the horizon was reduced to half of the default value, performance dropped substantially. This suggests that a short horizon prevents MPPI from sufficiently anticipating future system evolution, thereby limiting its ability to generate effective auxiliary control inputs. In our task, selecting good actions requires looking beyond immediate outcomes, and an insufficient horizon directly degrades performance.

In contrast, when the horizon was doubled, performance improved slightly, but the gain was marginal. This indicates that the default horizon was already long enough to capture most of the useful future information, and that extending it further yields diminishing returns.

\textit{Number of samples.}
We then evaluate the effect of the number of samples. When the number of samples was reduced to half, the performance decreased slightly. On the other hand, when it was increased to twice the default value, the overall performance remained largely unchanged. This suggests that the default number of samples was already sufficient to provide adequate exploration in the trajectory optimization process, and that performance had already begun to saturate around this range.

\paragraph{Performance of pure MPPI.}
Finally, we evaluate the performance of MPPI when used alone. In this experiment, MPPI was executed using the same reward model as that used in PPO-MPPI as the cost function, multiplied by $-1$ to convert reward maximization into cost minimization. This setting allows us to test whether MPPI alone can achieve strong performance when using the same reward model as PPO-MPPI.

The results show that pure MPPI performs significantly worse than PPO-MPPI, even though it has direct access to the same reward model. This finding suggests that knowing the reward function alone is not sufficient for effective control in this task. A key reason is that the reward function is not sufficiently smooth and dense from the perspective of finite-horizon trajectory optimization. When meaningful reward differences do not appear within the prediction horizon, MPPI struggles to distinguish good trajectories from bad ones and therefore cannot reliably generate effective control actions.

This result highlights the difference in roles between RL and MPPI. RL can learn useful long-term behavioral tendencies through repeated interaction and long-horizon credit assignment. MPPI, on the other hand, is effective at finite-horizon local optimization, but its performance can be limited when the reward is sparse, discontinuous, or difficult to exploit within a short prediction window. Therefore, in settings such as ours, it is more effective to combine RL for long-term guidance with MPPI for short-term control refinement.

\begin{figure}[t]
    \centering
    \includegraphics[width=0.5\columnwidth]{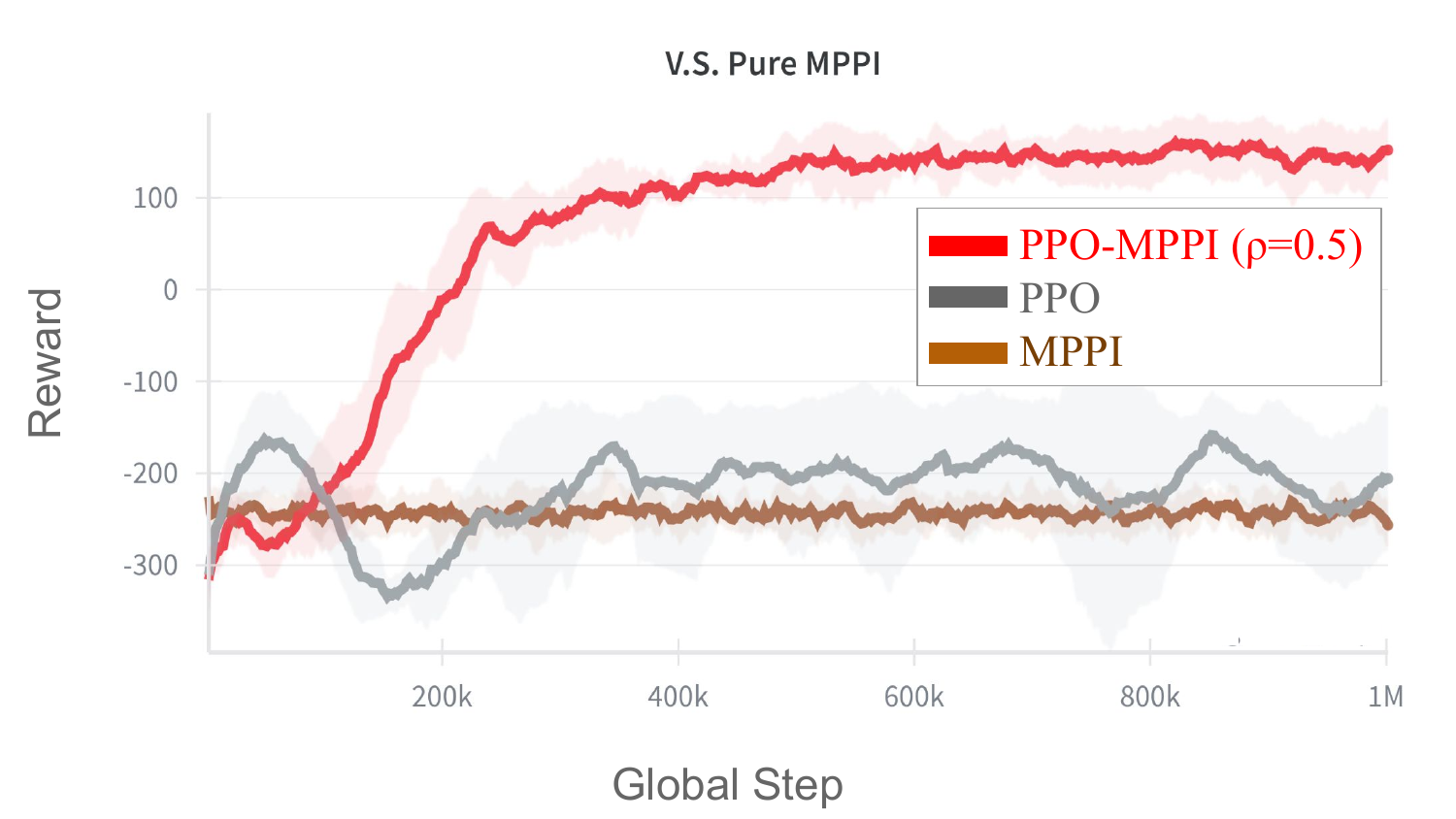}
    \caption{Performance comparison between PPO-MPPI and pure MPPI.}
    \label{fig:pure_mppi}
\end{figure}

\end{document}